\documentclass[letterpaper]{article} 
\usepackage{aaai25}  
\usepackage{times}  
\usepackage{helvet}  
\usepackage{courier}  
\usepackage[hyphens]{url}  
\usepackage{graphicx} 
\usepackage{natbib}  
\usepackage{caption} 
\usepackage{algorithm}
\usepackage{algorithmic}

\usepackage{graphicx}
\usepackage{booktabs}
\usepackage{lipsum,booktabs}
\usepackage{multirow}
\usepackage{threeparttable}
\usepackage{pifont}
\usepackage{todonotes}
\usepackage{amssymb}
\usepackage{colortbl}
\usepackage{bbding}
\usepackage{afterpage}
\usepackage{enumitem}
\usepackage{xcolor}

\definecolor{ForestGreen}{RGB}{34,139,34}
\definecolor{myyellow}{RGB}{181, 181, 27}
\newcommand{\cmark}{\ding{51}}%
\newcommand{\xmark}{\ding{55}}%
\newcommand{\greencheck}{{\color{ForestGreen}\cmark}}

\newcommand{\redcheck}{{\color{red}\xmark}}
\usepackage{adjustbox}
\usepackage{makecell}
\usepackage{tikz}
\newcommand{\blackcircle}[1]{%
\tikz[baseline=(char.base),baseline=-0.7ex]{\node[shape=circle,fill=black,text=white,inner sep=0.5pt,font=\scriptsize] (char) {#1};}%
}

\newcommand{\OurMethod}{\emph{MotionCraft}}
\newcommand{\Benchmark}{\emph{MC-Bench}}
\newcommand{\OurAttn}{\emph{MC-Attn}}

\usepackage{amsmath,amsfonts,bm}

\def\eqref#1{equation~\ref{#1}}

\def\1{\bm{1}}

\DeclareMathAlphabet{\mathsfit}{\encodingdefault}{\sfdefault}{m}{sl}
\SetMathAlphabet{\mathsfit}{bold}{\encodingdefault}{\sfdefault}{bx}{n}

\newcommand{\R}{\mathbb{R}}

\newcommand\blfootnote[1]{%
  \begingroup
  \renewcommand\thefootnote{}\footnote{#1}%
  \addtocounter{footnote}{-1}%
  \endgroup
}

\usepackage{newfloat}
\usepackage{listings}
\DeclareCaptionStyle{ruled}{labelfont=normalfont,labelsep=colon,strut=off} 
\floatstyle{ruled}
\newfloat{listing}{tb}{lst}{}
\floatname{listing}{Listing}
\title{MotionCraft: Crafting Whole-Body Motion with Plug-and-Play \\ Multimodal Controls}
\author{
    Yuxuan Bian\textsuperscript{\rm 1},
    Ailing Zeng\textsuperscript{\rm 2}\thanks{Corresponding authors.},
    Xuan Ju\textsuperscript{\rm 1},
    Xian Liu\textsuperscript{\rm 1},
    Zhaoyang Zhang\textsuperscript{\rm 1},
    Wei Liu\textsuperscript{\rm 2},
    Qiang Xu\textsuperscript{\rm 1}$^{*}$
}
\affiliations{
    \textsuperscript{\rm 1}The Chinese University of Hong Kong
    \textsuperscript{\rm 2}Tencent\\
    \url{https://cure-lab.github.io/MotionCraft}

}

\usepackage{bibentry}

\begin{document}

\twocolumn[{
\renewcommand\twocolumn[1][]{#1}
\maketitle
\vspace{-1.3cm}
\begin{center}
    \captionsetup{type=figure}
    \includegraphics[width=1.0\textwidth]{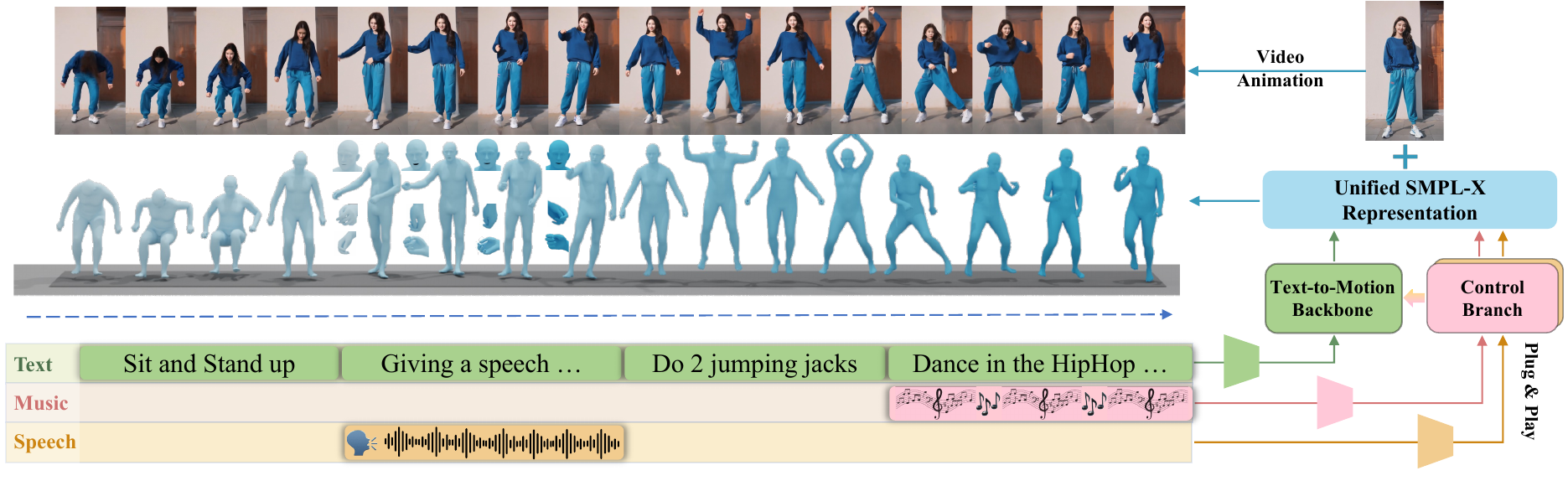}
    \vspace{-0.8cm}
    \captionof{figure}{
    We propose \OurMethod, a diffusion transformer that crafts whole-body motion with plug-and-play multimodal controls, encompassing robust motion generation abilities including Text-to-Motion, Speech-to-Gesture, and Music-to-Dance.
    }
\end{center}
}]

\begin{abstract}
\vspace{-0.2cm}
Whole-body multimodal motion generation, controlled by text, speech, or music, has numerous applications including video generation and character animation. 
However, employing a unified model to achieve various generation tasks with different condition modalities presents two main challenges: motion distribution drifts across different tasks (e.g., co-speech gestures and text-driven daily actions) and the complex optimization of mixed conditions with varying granularities (e.g., text and audio). Additionally, inconsistent motion formats across different tasks and datasets hinder effective training toward multimodal motion generation.
In this paper, we propose \OurMethod, a unified diffusion transformer that crafts whole-body motion with plug-and-play multimodal control.
Our framework employs a coarse-to-fine training strategy, starting with the first stage of text-to-motion semantic pre-training, followed by the second stage of multimodal low-level control adaptation to handle conditions of varying granularities. 
To effectively learn and transfer motion knowledge across different distributions, we design \OurAttn~for parallel modeling of static and dynamic human topology graphs.
To overcome the motion format inconsistency of existing benchmarks, we introduce \Benchmark, the first available multimodal whole-body motion generation benchmark based on the unified SMPL-X format.
Extensive experiments show that \OurMethod~achieves state-of-the-art performance on various standard motion generation tasks. 
\blfootnote{$^{*}$~Corresponding authors.}
\end{abstract}

%


\section{Introduction}
\label{sec:introduction}

Whole-body human motion generation with multimodal controls ~\citep{lmm,beats2,finedance}, which produces natural and coherent human movements based on multimodal conditions, 
has numerous applications, including human video generation~\citep{hu2024animate} and character animation~\citep{magicavatar}.

Recent advancements in single-conditioned human motion generation have made it possible to generate realistic human movements from a variety of control signals with varying granularities, including text descriptions~\citep{humanml3d,finemogen}, music clips~\citep{bailando,finedance}, and speech segments~\citep{beats2,diffsheg}. 
However, extending these capabilities to whole-body motion generation with multimodal control within a unified model introduces several significant challenges:

\begin{itemize}[leftmargin=*,itemsep=-0.1em]
    \item[\ding{224}] \textbf{\textit{Motion distribution drifts:}} Under different conditions, the motion distribution often varies significantly~\citep{lmm,mcm}. 
    In text-to-motion~(T2M), semantic text guidance mainly controls daily torso movements~\citep{humanml3d,motionx}, 
    while speech-to-gesture~(S2G) focuses on gestures and facial expressions under first-perspective audio~\citep{beats2,talkshow}. 
    Music-to-dance~(M2D) includes a more dynamic and variable correlation between the third-perspective music with limb movements~\citep{finedance}. Previous research usually focused on a single task to avoid the weak generative transferability posed by distribution drifts. 
   
    \item[\ding{224}] \textbf{\textit{Optimization challenges under mixed conditions:}} Current multimodal motion generation work compress diverse control signals--such as semantic text guidance, first-person speech, and third-person music--into a common latent space for mixed modeling. This includes transformer token embedding~\citep{ude2} and the feature space used in ImageBind~\citep{girdhar2023imagebind}. 
    However, this approach often leads to alignment issues across different modalities and introduces optimization challenges when learning conditions at different levels of granularity simultaneously~\citep{team2023gemini}.
    
    \item[\ding{224}] \textbf{\textit{Non-uniform whole-body motion format and evaluation:}} Finally, there are no high-quality multimodal whole-body human motion generation benchmarks with unified motion representation and evaluation pipelines.
\end{itemize}

In this work, we propose a unified motion diffusion transformer, \textbf{\textit{\OurMethod}}, that crafts whole-body motion with plug-and-play multimodal control, generating fine-grained text- and speech~(music)-aligned motions. 
It also supports generating motion with multiple conditions simultaneously, such as text combined with speech or music.

For effective learning of conditions with varying granularities, \OurMethod~employs a two-stage, coarse-to-fine multimodal generation framework. In the first stage, it captures high-level semantic motion generation abilities guided by coarse-grained text. In the second stage, control branches are added to the frozen backbone from the first stage, allowing the model to retain semantic generation capabilities while achieving fine-grained plug-and-play controls for specific low-level conditions~(speech or music) without the optimization confusion associated with mixed training.

To address motion distribution drifts across various generation scenarios, we analyze human motion kinematics and distribution using t-distributed stochastic neighbor embedding. We find that motion distributions corresponding to different control signals can be decomposed into static human topology structures and dynamic topology relationships, which are generalizable across different scenarios.
Different from the existing large language and vision model, the amount of motion data is still very small and unscalable. 
To model these human-centric spatiotemporal properties, we design \textbf{\textit{\OurAttn}}, where the spatial branch learns and transfers motion topology knowledge across different distributions by parallel modeling of static and dynamic human topology graphs, while the temporal branch captures the temporal relationships within the motion sequences. 

To overcome the inconsistent motion format limitation in existing benchmarks, such as Rot6D~\citep{humanml3d}, SMPL~\citep{smpl}, and SMPL-X~\citep{smplx}, we also introduce \textbf{\textit{\Benchmark}}, the first available multimodal motion generation benchmark based on the unified whole-body SMPL-X format, including data construction and evaluation pipelines. Extensive experiments demonstrate that \OurMethod~achieves competitive performance across various standard motion generation tasks, including text-to-motion, speech-to-gesture, and music-to-dance. Additionally, we provide comprehensive ablation studies, offering insights into model design and scaling effects for future multimodal whole-body motion generation models.

In summary, our contributions are as follows:
\begin{itemize}[leftmargin=*]
    \setlength\itemsep{0.1em}
     \item We propose \textbf{\textit{\OurMethod}}, a two-stage, coarse-to-fine multimodal motion generation framework that supports control signals at different granularities, enabling efficient plug-and-play multimodal motion generation.
     
    \item We design \textbf{\textit{\OurAttn}}, the first attempt to achieve modeling of static and dynamic human topology against motion distribution drifts in multimodal motion generation.
    \item We create \textbf{\textit{\Benchmark}}, the first publicly available multimodal whole-body motion generation benchmark with a unified whole-body motion representation SMPL-X.
\end{itemize}

\begin{table*}[!htbp]
\vspace{-0.5cm}
    \centering
    \small
    \caption{
    \textbf{Comparison of \OurMethod~with previous motion generation methods.} \OurMethod~jointly models the static human skeleton structure and dynamic human topology relationships to achieve flexible motion knowledge transfer across various whole-body generation scenarios, supporting plug-and-play with any new control signal modality.
    }
    \vspace{-0.2cm}
    \scalebox{0.75}{
    \renewcommand\arraystretch{0.8}
\setlength{\tabcolsep}{0.9mm}{
    \begin{tabular}{cccccccccc}
\toprule
Model                       & Text2Motion  & Music2Dance & Speech2Gesture & Static Body Prior & Dynamic Body Adaption & Whole Body & Unified Representation &  Plug-and-Play \\ 
\midrule
FineMoGen~\citep{finemogen} & \greencheck & \redcheck & \redcheck & \greencheck & \redcheck & \redcheck & \redcheck & \redcheck \\
HumanTomato~\citep{humantomato} & \greencheck & \redcheck & \redcheck & \greencheck & \redcheck & \greencheck & \redcheck & \redcheck\\
FineDance~\citep{finedance} & \redcheck & \greencheck & \redcheck  & \redcheck & \redcheck & \redcheck & \redcheck & \redcheck\\
Bailando~\citep{bailando} & \redcheck & \greencheck & \redcheck  & \greencheck & \redcheck & \redcheck & \redcheck & \redcheck\\
EMAGE~\citep{beats2} & \redcheck & \redcheck & \greencheck & \greencheck & \redcheck & \greencheck & \redcheck & \redcheck\\
TalkShow~\citep{talkshow} & \redcheck & \redcheck & \greencheck & \greencheck & \redcheck & \greencheck & \redcheck & \redcheck\\
MCM~\citep{mcm} & \greencheck & \greencheck & \greencheck & \redcheck & \redcheck & \redcheck & \redcheck & \greencheck\\
Motion-Verse~\citep{lmm} & \greencheck & \greencheck & \greencheck & \redcheck & \greencheck & \greencheck & \redcheck & \redcheck\\
\midrule
\OurMethod            &   \greencheck  &    \greencheck &  \greencheck  &  \greencheck  &  \greencheck  &  \greencheck  &  \greencheck  &  \greencheck \\ 
\bottomrule
\end{tabular}}}
\vspace{-0.4cm}
    \label{tab:compare_other_methods}
\end{table*}

\section{Related Work}
\label{sec:related_work}

\subsection{Human Motion Generation Models}
Conditioned human motion generation models have made significant progress, including text-to-motion~(T2M)~\citep{mdm,t2mgpt,liu2023plan,motiondiffuse,finemogen,liang2024omg}, speech-to-gesture~(S2G)~\citep{talkshow,diffsheg,liu2022learning}, and music-to-dance~(M2D)~\citep{finedance,edge,bailando}.
Recently, increasing attention has been paid to multimodal motion generation~\citep{mcm,lmm,m3gpt}. 
$M^{3}$-GPT~\citep{m3gpt} injects quantized condition tokens into the vocabulary of large language models to achieve motion understanding and generation, but it overlooks the modeling of human topology priors.
Motion-Verse~\citep{lmm} incorporates dynamic attention to assess relationships among body parts but fails to capture the overall static human topology, leading to limited generalization power and increased optimization complexity. 
Furthermore, it employs mixed training across all conditions based on ImageBind~\citep{girdhar2023imagebind}, which creates optimization challenges when learning conditions of varying granularities simultaneously and needs retraining for new control signals.
MCM~\citep{mcm} attempts to address the optimization confusion of mixed training based on the ControlNet~\citep{zhang2023adding} architecture, but it neglects any modeling of human topology structure, resulting in poor generalization across generation scenarios. 
Compared to previous methods in Tab.~\ref{tab:compare_other_methods}, \OurMethod~generates whole-body motion under varying control signals with plug-and-play capability by using \OurAttn~to capture static human topology and domain-specific dynamic skeleton relationships, incorporating control branches, and employing a coarse-to-fine training strategy.

\subsection{Human Motion Generation Benchmarks}
Various conditioned human motion generation benchmarks have been constructed in recent years.
For T2M, researchers have curated datasets encompassing action categories~\citep{chung2021haa500,trivedi2021ntu}, sequential action labels~\citep{zhang2022egobody,guo2020action2motion}, and arbitrary natural language descriptions~\citep{motionx,humanml3d,flag3d_cvpr}.
For M2D, 
AIST++~\citep{aistpp} reconstructs 5 hours of dance based on SMPL~\citep{smpl} format from videos. 
Finedance~\citep{finedance} collects dances of 14.6 hours across 22 genres and supplements the dataset with detailed gestures using the SMPL-H~\citep{smplx} format.
For S2G datasets~\citep{beats2,beats,talkshow}, BEAT2~\citep{beats2} and BEAT~\citep{beats} have emerged as the most popular benchmarks, celebrated for their diverse range of motion and extensive data volume. BEAT2, built upon BEAT, utilizes SMPL-X and FLAME~\citep{kim2023flame} to achieve higher-quality unified mesh-level data.
Despite these developments, no publicly available benchmark supports unified representation for multimodal whole-body motion generation.

\section{Motivation}
\label{sec:preliminaries_and_motivation}

The key challenge in achieving whole-body human motion generation with multimodal controls is addressing motion distribution drifts across different generation scenarios~\citep{lmm} and the efficient learning of control signals at varying granularities~\citep{mcm}.

\begin{figure}[]
    \centering
    \vspace{-0.2cm}
    \includegraphics[width=1.0\linewidth]{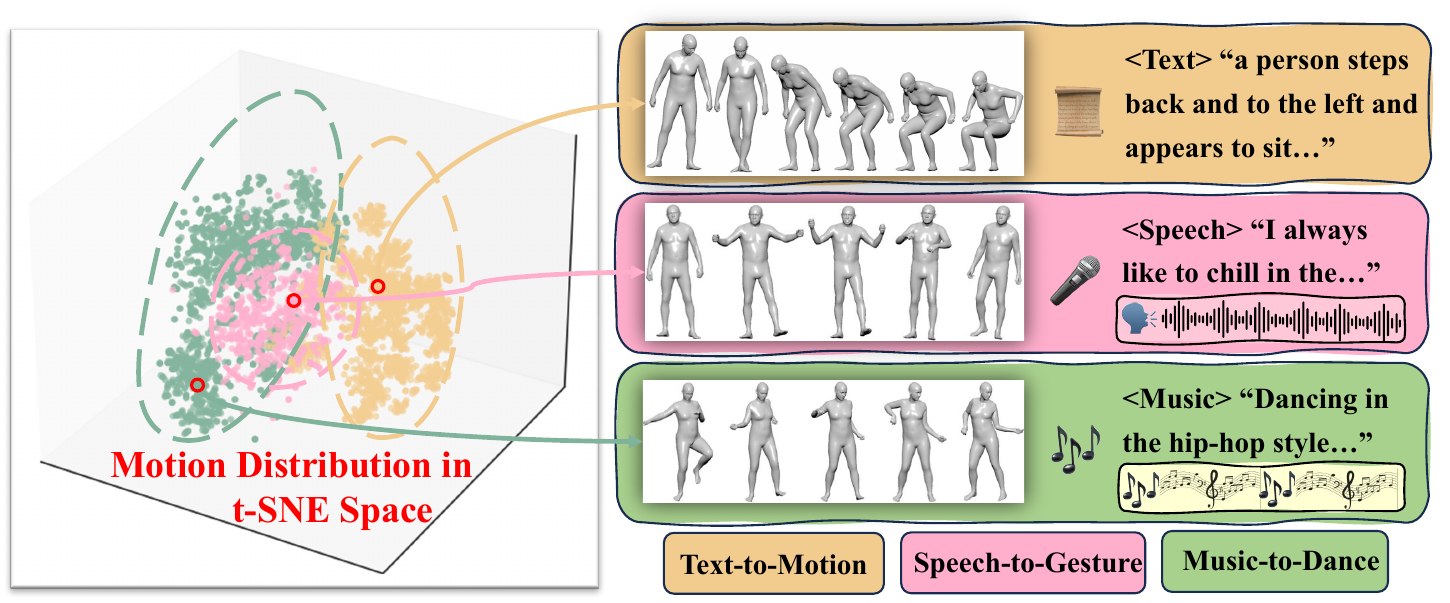}
    \vspace{-0.5cm}
    \caption{
    \textbf{The t-SNE latent space of motion in different generation tasks.} It illustrates the motion distribution drifts across different generation scenarios. 
    }
    \label{fig:motivation}
    \vspace{-0.6cm}
\end{figure}

\textbf{\textit{Motion distribution drifts solution.}} 
Current motion generation models mainly focus on scenarios with a single condition since they struggle to handle the noticeable motion distribution drifts across different scenarios~\citep{ude2}. 
For instance, as shown in Fig.~\ref{fig:motivation}, T2M primarily involves everyday torso movements, S2G includes complex hand gestures, rich facial expressions, and almost stationary lower limbs, while M2D emphasizes varied and extensive limb movements with limited hand movements.
However, many human-centric studies~\citep{zeng2021learning,ma2022pretrained} have confirmed that representing the human skeletal topology as a directed weighted graph, with different body parts as vertices, can introduce kinematic priors in complex motion modeling, thereby improving generalizability under distribution shifts. 
Additionally, based on human kinematic~\citep{smpl,smplx}, it is natural to decompose the human skeleton into a combination of static and dynamic topologies. 
For instance, in any scenario, the root vertex (hip) always significantly influences its child vertices (lower limbs or upper arms), with symmetrical interactions between pairs of arms.
However, in S2G, the correlation between the limbs and other body parts weakens, while the linking weight between hands and facial expressions strengthens.
Therefore, modeling both dynamic and static topology graphs can efficiently generalize motion knowledge across different generation tasks, even with limited data and significant distribution drifts.

\textbf{\textit{Efficient learning of conditions at varying granularities.}}
Different motion generation scenarios correspond to conditions at varying granularities. 
For instance, text guidance typically provides sequence-level coarse-grained semantic control, while speech and music focus more on per-frame low-level control~\citep{beats2,finedance}. 
Mixed learning of all conditions within a single space leads to inevitable modality alignment loss and fails to decouple the learning process for each granularity~\citep{zhang2023adding,mcm}, causing optimization confusion. 
Motivated by other vision generation paradigms in the image/video domain, including StableDiffusion~\citep{sd} and Sora~\citep{sora}, decoupling the generation under different conditions and using T2M as a basic pre-training task can build robust generative abilities for following multi-condition generation, resulting in more efficient and fine-grained multimodal control generation.

\begin{figure*}[!ht]
    \centering
    \vspace{-0.4cm}
    \includegraphics[width=\linewidth]{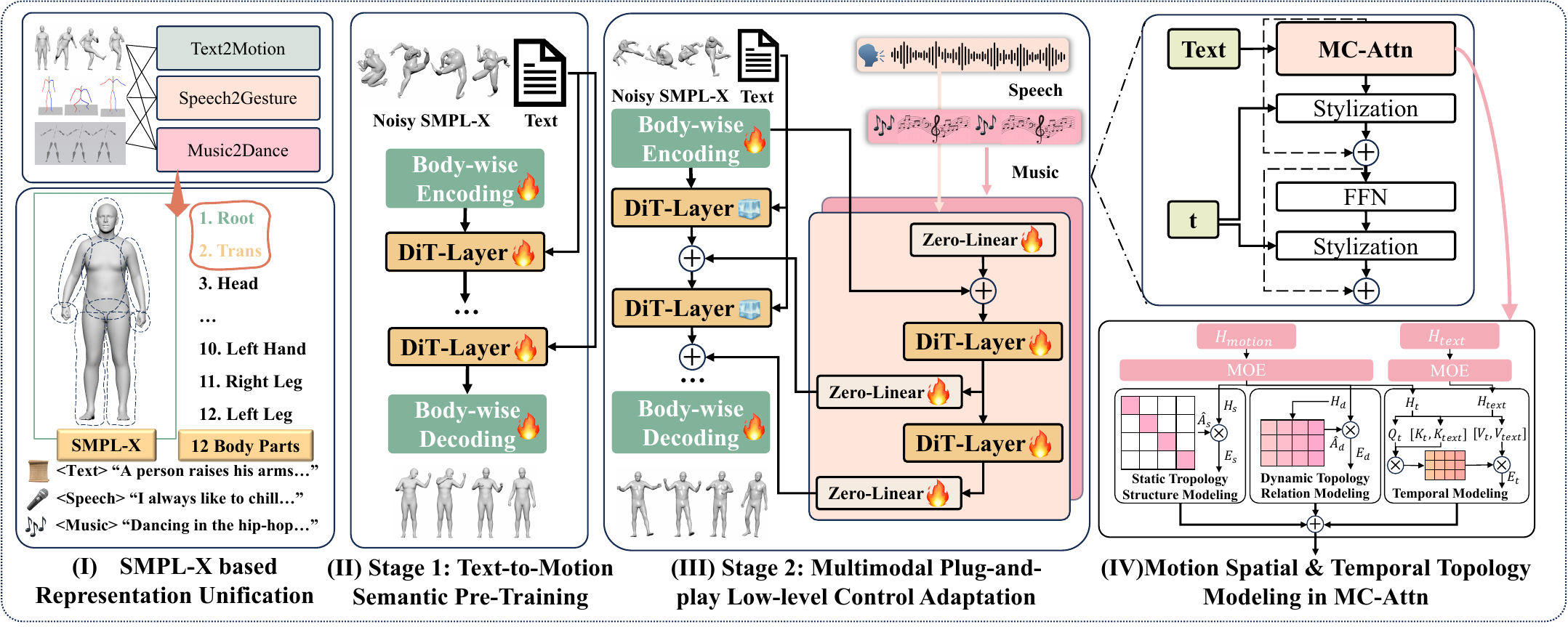}
    \vspace{-0.6cm}
    \caption{
    \textbf{Architecture of \OurMethod.} \OurMethod~is a transformer-based diffusion model. 
    In the first stage, \OurMethod~uses text as a semantic control guide to learn coarse-grained cross-scenario motion knowledge across multiple datasets; in the second stage, \OurMethod~freezes the backbone while adding a plug-and-play control branch to learn the different low-level control signals.
    The core of \OurMethod~is \OurAttn, which optimizes the representation of motion token sequences by capturing the spatial properties of static and dynamic human topology graphs and learning temporal relationships in parallel.
    }
    \vspace{-0.5cm}
    \label{fig:method}
\end{figure*}

\section{Proposed Method}
\label{sec:method}

\subsection{\OurMethod~Framework}
\label{sec:framework}
The overview of \OurMethod~is described in Fig. \ref{fig:method}. 
Aimed at decoupling the conditioned generation learning at varying granularities, we adopt a two-branch architecture consisting of a main text-to-motion branch and a plug-and-play low-level control branch, along with a two-stage coarse-to-fine training strategy to efficiently grasp the motion topology knowledge across different scenarios with various control signal modalities. 
Both branches use a motion diffusion transformer specifically designed with \OurAttn~to capture both static and dynamic motion topology properties.

\textbf{Stage~\blackcircle{1}~Text-to-Motion Semantic Pre-training.}
The main branch $f_{m}(\cdot)$ is optimized in Stage I, text-to-motion semantic pre-training, using text-to-motion paired data collected from diverse scenarios in \Benchmark. 
We choose text as the shared condition among various unimodal datasets, allowing \OurMethod~to acquire sequence-level generation and coarse-grained text-guidance following abilities between text $\mathbf{H}_{text} \in \R^{B \times F_{t} \times D_{t}}$ and motion $\mathbf{H}_{motion} \in \R^{B \times F_{m} \times D_{m}}$. 
Overall, text guidance pre-training in diverse generation scenarios helps follow fine-grained controls of other low-level conditions in Stage II.

\textbf{Stage~\blackcircle{2}~Multimodal Low-level Control Adaptation.}
During the low-level control adaptation fine-tuning stage, we aim to model the correlation between various condition signals $\mathbf{H}_{c} \in \R^{B \times T_{c} \times D_{c}}$ and motion sequences $\mathbf{H}_{motion} \in \R^{B \times F_{m} \times D_{m}}$. 
All main branch parameters $f_{m}(\cdot)$ are frozen to maintain their coarse-grained motion generation and semantic text guidance following abilities.
A copy of the main branch parameters $\hat{f_{m}}(\cdot)$ is then used to initialize the control branch, connecting them with a zero-initialized linear layer $\mathbf{W}_{p} \in \R^{D_{m} \times D_{m}}$ to prevent early training noise from causing collapse. 
Then the condition signals (speech, music, or other low-level control signals) are fed into the control branch, where a position mask $\mathbf{M}_{c} \in \{0, 1\}^{F_{m}}$ aligns the condition signals to the motion sequence length $F_{m}$, setting zeros for the $F_{m} - T_{c}$ missing frames in the original control signal sequence. 
The output of each control branch layer is directly added to the corresponding main branch layer input through the zero bridge linear, allowing new control signals to guide frame-level human motion generation.

\subsection{\OurAttn~Design}
\label{sec_architecture}

The core of \OurMethod~is \OurAttn, which parallel captures the static and dynamic human topology graphs, thereby enhancing the transferability of motion topology knowledge across diverse generation scenarios against non-neglectable distribution drifts.
\OurAttn~has three key components: a static-skeleton graph learner and a dynamic-topology relationship graph learner for parallel modeling motion spatial properties, and temporal attention for modeling the frame-level dynamics of each body part over time.
The three modules share the same motion representation input $\mathbf{H}_{m} \in \R^{B \times F_{m} \times D_{m}}$, the output of the last \OurAttn~layer refined further by a MOE\citep{moe}.

For the static-skeleton graph learner, the process begins by constructing the $N_{b}$ graph vertex representation $\mathbf{H}_{s} \in \mathbb{R}^{B \times F_{m} \times N_{b} \times D_{b}}$, followed by initializing a diagonal unit matrix $\mathbf{A}_{s} \in \mathbb{R}^{N_{b} \times N_{b}}$ as the adjacency matrix for the initial static topology graph $\mathcal{G}_{s}$, where each body part is connected only to itself to avoid training collapse from random connections. 
Through optimization, $\hat{\mathbf{A}}_{s}$ captures the static, input-independent human topology, enabling the model to quickly grasp fundamental human structure for new scenarios, even with limited data. The module outputs $\mathbf{E}_{s} = \hat{\mathbf{A}}_{s} \cdot \mathbf{H}_{s}$.

While static human topology graphs capture the basic structure and facilitate quick convergence to new distributions, they do not adapt dynamically to new contexts, potentially causing underfitting in evolving scenarios~\citep{lmm}. 
To address this, we introduce a dynamic-topology relationship graph learner that models dynamic distribution features and adjusts to distribution drifts based on control signals, complementing the static topology structure.  
Specifically, the dynamic-topology relationship graph learner represents each body part as a dynamic graph vertex $\mathbf{H}_{d} \in \R^{B \times F_{m} \times N_{b} \times D_{b}}$, using attention scores $\mathbf{A}_{d} \in \R^{B \times F_{m} \times N_{b} \times N_{b}}$ as edge weights in the dynamic human topology graph~$\mathcal{G}_{d}$, thereby enhancing the model’s ability to adapt its spatial structure beyond the static-skeleton learner. The final output is $\mathbf{E}_{d} = \mathbf{A}_{d} \cdot \mathbf{H}_{d}$.

Various studies have shown that basic attention is sufficient for modeling temporal relationships~\citep{nie2022time,bianmulti}. Therefore, we chose to use each body part as a unit $\mathbf{H}_{t} \in \R^{B \cdot N_{b} \times F_{m} \times D_{b}}$ and measure the temporal relationships between frames based on attention~\citep{vaswani2017attention}.
Considering that external textual control signals are mostly sequential instructions in the temporal dimension, text information is also modeled here to produce the output, $\hat{\mathbf{E}_{t}} = Softmax(\mathbf{Q}_{H_{t}}\cdot [\mathbf{K}_{H_{t}}^{T}, \mathbf{K}_{H_{text}}^{T}]/\sqrt{D_{b}})\cdot [\mathbf{V}_{H_{t}}^{T}, \mathbf{V}_{H_{text}}^{T}]$, where $\mathbf{Q}_{H_{t}}=\mathbf{W}^{Q_{H_{t}}}\mathbf{H}_{t}$, $\mathbf{K}_{H_{t}}=\mathbf{W}^{K_{H_{t}}}\mathbf{H}_{t}$,
$\mathbf{K}_{H_{text}}=\mathbf{W}^{K_{H_{text}}}\mathbf{H}_{text}$,
$\mathbf{V}_{H_{t}}=\mathbf{W}^{V_{H_{t}}}\mathbf{H}_{t}$, $\mathbf{V}_{H_{text}}=\mathbf{W}^{V_{H_{text}}}\mathbf{H}_{text}$,
and $[,]$ denotes concat operation.
Other sequential control modalities, such as speech and music, are modeled in the control branch. 
The final output $\mathbf{E} = \mathbf{E}_{s} + \mathbf{E}_{d} + \mathbf{E}_{t}$ of \OurAttn~combines the spatiotemporal representations of the human skeleton and the temporal dynamics of each body part.

\begin{figure*}[!ht]
    \centering
    \vspace{-0.5cm}
    \includegraphics[width=1.00\linewidth]{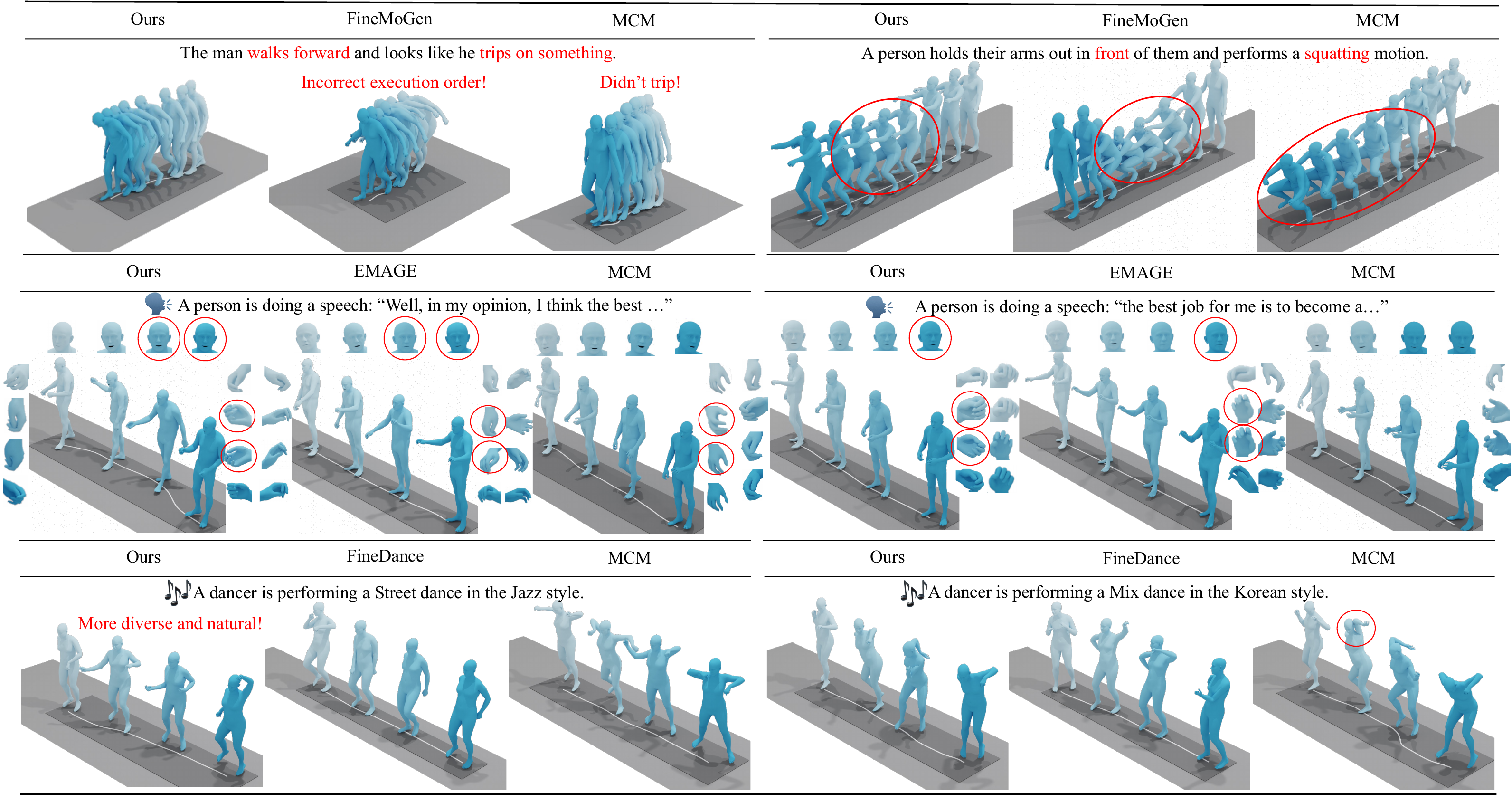}
    \vspace{-0.7cm}
    \caption{
    The qualitative results of \OurMethod~and other state-of-the-art baselines on three representative tasks, text-to-motion, speech-to-gesture, and music-to-dance. More detailed visualization comparisons are in our supplementary.
    }
    \vspace{-0.2cm}
    \label{fig:visual}
\end{figure*}

\subsection{\Benchmark~Construction}
\label{sec:bench}
To prevent the information loss when aligning different motion formats, we select HumanML3D~\citep{humanml3d} in SMPL format for T2M, FineDance~\citep{finedance} in SMPL-H Rot-6D format for M2D, and BEAT2~\citep{beats2} in SMPL-X format for S2G from public datasets, as they are the most representative unimodal datasets in their respective areas. To enable whole-body multimodal control of human motion generation, we converted all data to the SMPL-X format. Key operations include filling in missing facial information in HumanML3D and FineDance with average expressions and converting FineDance from SMPL-H Rot-6D format to axis-angle representation for efficient alignment with SMPL-X parameters and minimal alignment errors compared to the official body-retargeting method. 
We then pre-train a motion encoder and a text encoder by aligning text and motion contrastively with a retrieval optimization goal~\citep{humantomato} for a unified evaluation of the SMPL-X motion representation.
For FineDance and BEAT2, which lack corresponding textual information, we generate pseudo-captions such as "A dancer is performing a street dance in the Jazz style to the rhythm of the wildfire" and "A person is giving a speech, and the content is ...".
\section{Experiments}
\label{sec:experiments}

\subsection{Implementation Details}
We designed two model variants for the first stage of Text-to-Motion backbone training, \OurMethod-Basic and \OurMethod-Mix, which were trained on the HumanML3D subset in \Benchmark~and the entire \Benchmark, respectively. 
In the second stage, we used BEAT2~\citep{beats2}, a large dataset for speech gesture synthesis, and FineDance~\citep{finedance}, a high-quality choreography dataset, to train control branches for Speech-to-Gesture and Music-to-Dance. 
\OurMethod-Basic and \OurMethod-Mix share the same 4-layer transformer backbone configuration, dividing the body topology into 12 parts, each with a body-part hidden encoding dimension of 64. 
\Benchmark~used a unified whole-body motion format SMPL-X~\citep{smplx} in the form of axis-angle, instead of the joint positions or 6D rotation. Thus we retrained the motion and text encoder based on SMPL-X using OpenTMR~\citep{humantomato} for evaluation.

\begin{table*}[!ht]
    \centering
    \resizebox{1.00\linewidth}{!}{
        \begin{tabular}{c|ccc|ccc}
            \toprule
            \multirow{2}{*}{Method} & \multicolumn{3}{c|}{R Precision} & \multirow{2}{*}{FID $\downarrow$} & \multirow{2}{*}{Div $\uparrow$} & \multirow{2}{*}{MM Dist$\downarrow$}  \\
            \cline{2-4}
            & Top-1 $\uparrow$ & Top-2 $\uparrow$ & Top-3 $\uparrow$ \\
            \hline
            GT &$0.663^{\pm{0.006}}$ & $0.807^{\pm{0.002}}$ & $0.864^{\pm{0.002}}$ &  $0.000^{\pm{0.000}}$ & $36.423^{\pm{0.183}}$ & $15.567^{\pm{0.036}}$  \\
            \hline
            T2M-GPT\cite{t2mgpt} & $0.529^{\pm{0.004}}$ & $0.652^{\pm{0.003}}$ & $0.732^{\pm{0.003}}$ &  $10.457^{\pm{0.108}}$ & $36.114^{\pm{0.098}}$ & $17.029^{\pm{0.039}}$ \\
            MDM\cite{mdm} & $0.383^{\pm{0.010}}$ & $0.527^{\pm{0.012}}$ & $0.604^{\pm{0.009}}$ &  $18.671^{\pm{0.370}}$ & $36.156^{\pm{0.103}}$ & $18.785^{\pm{0.054}}$ \\
            MotionDiffuse\cite{motiondiffuse} & $0.525^{\pm{0.004}}$ & $0.675^{\pm{0.009}}$ & $0.743^{\pm{0.009}}$ &  $9.982^{\pm{0.379}}$ & $36.187^{\pm{0.160}}$ & $17.314^{\pm{0.066}}$ \\
            FineMoGen\cite{finemogen} & $0.565^{\pm{0.001}}$ & $0.710^{\pm{0.004}}$ & $0.775^{\pm{0.004}}$ &  \cellcolor{yellow!15}$7.323^{\pm{0.143}}$ & $36.324^{\pm{0.069}}$ & $16.679^{\pm{0.029}}$ \\
            MCM\cite{mcm} & $0.407^{\pm{0.002}}$ & $0.559^{\pm{0.003}}$ & $0.636^{\pm{0.001}}$ &  $15.540^{\pm{0.443}}$ & $35.813^{\pm{0.137}}$ & $18.673^{\pm{0.029}}$ \\
            
            \hline
            \OurMethod-Basic & 
            \cellcolor{yellow!15}$0.590^{\pm{0.003}}$ & \cellcolor{yellow!15}$0.743^{\pm{0.002}}$ & \cellcolor{yellow!15}$0.804^{\pm{0.004}}$ & $8.477^{\pm{0.102}}$ & \cellcolor{yellow!15}$36.210^{\pm{0.089}}$ & \cellcolor{red!15}$16.252 ^{\pm{0.035}}$ \\
            \OurMethod-Mix & 
            \cellcolor{red!15}$0.600^{\pm{0.003}}$ & \cellcolor{red!15}$0.747^{\pm{0.004}}$ & \cellcolor{red!15}$0.812^{\pm{0.006}}$ & \cellcolor{red!15}$6.707^{\pm{0.081}}$ & \cellcolor{red!15}$36.419^{\pm{0.047}}$ & \cellcolor{yellow!15}$16.334 ^{\pm{0.059}}$ \\
            \bottomrule
        \end{tabular}
    }
    \vspace{-0.2cm}
    \caption{\textbf{Results of Text-to-Motion in HumanML3D of \Benchmark.} 
    We compare the results of text-to-motion between ours and the SOTA methods. 
    $\colorbox{red!15}{\rm Red background indicates best results}, \colorbox{yellow!15}{\rm yellow background indicates second best results}$.}
    \label{tab:t2m_res_2}
    \vspace{-0.4cm}
\end{table*}
\subsection{Evaluation Metrics}
\paragraph{Text-to-Motion.}
We use Fr\'echet Inception Distance (\textbf{FID}) to measure the distribution distance between generated motion and the ground truth, and diversity (\textbf{Div}) to measure the average pairwise Euclidean distance among random pairs of generated motion. Furthermore, we use \textbf{R-Precision} to measure how often the top-k closest motions to their corresponding captions are achieved within a 32-sample batch. Finally, we employ Multi-Modal Distance (\textbf{MM Dist}) to quantify the average Euclidean distance between motion representations and their corresponding text features.

\paragraph{Speech-to-Gesture.}
We use $FID_{H}$, $FID_{B}$, and Div for quality and diversity measurement. $FID_{H}$ represents the difference between the hand motion distribution and the ground truth gesture distribution, while $FID_{B}$ focuses on the distance between the distributions of whole-body motion.
Moreover, 
we use the Beat Alignment Score~\citep{aistpp} to measure the alignment between the motion and speech beats and employ L2 Loss to measure the difference between generated and real expressions.

\paragraph{Music-to-Dance.}
Similar to Speech-to-Gesture, we use $FID_{H}$, $FID_{B}$, and Div to measure the quality of music-to-motion generation for hand and whole-body movements, as well as the diversity of the generated motions.

\subsection{Quantitative and Qualitative Results}
We evaluate \OurMethod~on three representative tasks: \ding{172}~Text-to-Motion, \ding{173}~Speech-to-Gesture, and \ding{174}~Music-to-Dance, analysing both quantitative and qualitative results.\footnote{Missing expressions are filled with zero for generated motion and ground truth to avoid affecting the evaluation results.}
More visualization comparisons are in our supplementary.

\begin{table*}
    \centering
    \vspace{-0.2cm}
    \resizebox{1.0\linewidth}{!}{
        \begin{tabular}{c|cc|ccc|c|ccc}
            \toprule
            {\textbf{S2G-Method}} & {$FID_{H}$ $\downarrow$} & {$FID_{B}$ $\downarrow$} & {Face L2 Loss $\downarrow$} & {Beat Align Score $\uparrow$} & {Div $\uparrow$} & {\textbf{M2D-Method}} & {$FID_{H}$ $\downarrow$} & {$FID_{B}$ $\downarrow$} & {Div $\uparrow$} \\
            \hline
            
            Talkshow & $26.713$ & $74.824$ & \cellcolor{yellow!15}$7.791$ & $6.947$ & \cellcolor{red!15}$13.472$ & Edge & $93.430$ & $108.507$ & $13.471$ \\ 
            EMAGE & $39.094$ & $90.762$ & \cellcolor{red!15}$7.680$ & $7.727$ & $13.065$ & Finedance & $10.747$ & \cellcolor{yellow!15}$72.229$ & $13.813$ \\ 
            MCM & $23.946$ & $71.241$ & $16.983$ & $7.993$ & \cellcolor{yellow!15}$13.167$ & MCM & $4.717$ & $78.577$ & $14.890$ \\ 
            \hline
            \OurMethod-Basic & \cellcolor{yellow!15}$18.486$ & \cellcolor{yellow!15}$27.023$ & $10.097$ & \cellcolor{yellow!15}$8.098$ & $10.334$ & \OurMethod-Basic & \cellcolor{yellow!15}$3.858$ & $76.248$ & \cellcolor{yellow!15}$16.667$\\ 
            \OurMethod-Mix & \cellcolor{red!15}$12.882$ & \cellcolor{red!15}$25.187$ & $8.906$ & \cellcolor{red!15}$8.226$ & $12.595$ & \OurMethod-Mix & \cellcolor{red!15}2.849  &  \cellcolor{red!15}67.159 & \cellcolor{red!15}18.483\\ 
            \bottomrule
        \end{tabular}
    }
    \vspace{-0.2cm}
    \caption{\textbf{Results of Speech-to-Gesture in BEAT2 and Music-to-Dance in FineDance of \Benchmark.} We respectively evaluate the $FID_{H}$ and $FID_{B}$, Face L2 Loss$\times 10^{-8}$, Beat Align Score$\times 10^{-1}$, and diversity for S2G and the $FID_{H}$, $FID_{B}$, and the diversity for M2D. $\colorbox{red!15}{\rm Red background indicates best results}, \colorbox{yellow!15}{\rm yellow background indicates second best results}$.}
    \label{tab:s2g_m2d_res_1}
\end{table*}

\paragraph{Comparison on Text-to-Motion Generation.}

In the text-to-motion task, we compare \OurMethod~with current SOTA baselines~\citep{finemogen,mcm,lmm,motiondiffuse,mdm,t2mgpt} in two benchmarks: the HumanML3D subset with whole-body format SMPL-X of \Benchmark~in Tab. \ref{tab:t2m_res_2} and the original HumanML3D~\citep{humanml3d} with the tensor-only format~(The results are in supplementary due to page limit). 
In both benchmarks, \OurMethod~achieved better text-guided generation capability, diversity, and motion generation quality. 
Notably, in the HumanML3D subset of \Benchmark, the inadequate evaluation abilities in the original HumanML3D benchmark with torso-only representation were significantly improved, providing a more comprehensive and objective comparison. This is because the whole-body SMPL-X representation requires the model to generate the torso movements, gestures, and expressions rather than the only torso. 
Additionally, we found that \OurMethod-Mix trained on the \Benchmark~has a significant advantage over \OurMethod-Basic. This is because \OurMethod-Mix~can efficiently transfer human topology knowledge against distribution drifts in various generation scenarios.
Visualization is in Fig. \ref{fig:visual}, and \OurMethod~can follow diverse textual descriptions with fine-grained control.

\begin{table*}
    \centering
    \vspace{-0.1cm}
    \resizebox{1.0\linewidth}{!}{
        \begin{tabular}{cc|ccccc|ccccc|ccc}
            \toprule
            \multicolumn{2}{c|}{\textbf{Method}} & \multicolumn{5}{c|}{\textbf{HumanML3D~(Text-to-Motion)}} & \multicolumn{5}{c|}
            {\textbf{BEAT2~(Speech-to-Gesture)}}  & \multicolumn{3}{c}{\textbf{Finedance~(Music-to-Dance)}} \\
            \midrule
            {Dynamic-Spatial} & 
            {Static-Spatial} & 
            Top-1 $\uparrow$ & Top-2 $\uparrow$ & Top-3 $\uparrow$ & {FID $\downarrow$} & {Div $\uparrow$} & {$FID_{H}$ $\downarrow$} & {$FID_{B}$ $\downarrow$} & {Face L2 $\downarrow$} & {Beat Align Score $\uparrow$} & {Div $\uparrow$} & {$FID_{H}$ $\downarrow$} & {$FID_{B}$ $\downarrow$} & {Div $\uparrow$} \\
             \midrule
             \redcheck & \redcheck & 0.583 & 0.729 & 0.794 & 8.911 & 35.954 & 15.587 & 31.839 & 12.448 & 7.908 & 11.752 & 7.088 & 150.733 & 17.984 \\
             \redcheck & \greencheck & 0.557 & 0.706 & 0.772 & 9.041 & 36.101 & 12.929 & 27.928 & 12.287 & 8.077 & 12.230 & 5.104  & 112.186 & \cellcolor{red!15}18.503 \\
             \greencheck & \redcheck & 0.582 & 0.732 & 0.798 & 8.455 & \cellcolor{red!15}36.241 & 15.517 & 28.631 & 12.544 & 7.708 & 11.313 & 4.972 & 102.103 & 16.385 \\
             \greencheck & \greencheck & \cellcolor{red!15}0.600 & \cellcolor{red!15}0.747 & \cellcolor{red!15}0.812 & \cellcolor{red!15}6.707 & 36.419 & \cellcolor{red!15}12.882 & \cellcolor{red!15}25.187 & \cellcolor{red!15}8.906 & \cellcolor{red!15}8.226 &  \cellcolor{red!15}12.595& \cellcolor{red!15}2.849 & \cellcolor{red!15}67.159 &  {18.483} \\
             \midrule
             \multicolumn{2}{c|}{\OurMethod-Tiny-(4, 64, 77M)} &  0.600 & 0.747 & 0.812 & 6.707 & \cellcolor{red!15}36.419 &  \cellcolor{red!15}12.882 & 25.187 & 8.906 & \cellcolor{red!15}8.226 &  \cellcolor{red!15}12.595 &  2.849 & 67.159 &  \cellcolor{red!15}{18.483} \\
             \multicolumn{2}{c|}{\OurMethod-Small-(4, 128, 130M)} &  \cellcolor{red!15}{0.653} & \cellcolor{red!15}{0.794} & \cellcolor{red!15}{0.847} & \cellcolor{red!15}{5.593} & {36.264} &   {15.346} & 27.140 & 8.322 & 8.023 & 11.906 &  \cellcolor{red!15}{2.370}  & \cellcolor{red!15}{59.471}  & 17.036 \\
             \multicolumn{2}{c|}{\OurMethod-Small-(8, 64, 145M)} &  0.635 & 0.779 & 0.802 & 6.193 & 36.311 &  15.702 & 28.094 & 8.589 &  8.031 &  11.824 &  3.749  & 66.958  & 16.478   \\
             \multicolumn{2}{c|}{\OurMethod-Medium-(8, 128, 250M)} & 0.647 & 0.785 & 0.854 & 5.670 & 36.384 & 14.937 & \cellcolor{red!15}23.498 & \cellcolor{red!15}8.125 & 8.089 & 10.962 & 3.904 & 75.412 & 16.507 \\
             \multicolumn{2}{c|}{\OurMethod-Large-(16, 128, 478M)} & 0.604 & 0.744 & 0.809 & 7.872 & 36.169 & 15.964 & 27.476 & 9.036 & 7.969 & 10.625 & 4.837 & 77.341 & 16.426 \\
             \bottomrule
        \end{tabular}
    }
    \vspace{-0.2cm}
    \caption{\textbf{Ablation Study.} 
    \textbf{(a) Ablation on model design~(Upper half).} The results suggest that jointly modeling dynamic and static human skeleton topologies significantly improves performance since this provides robust topology knowledge against distribution drifts. 
    \textbf{(b) Ablation on scaling up impacts~(Lower half).}
    We design four scaling model variants, where **-($a$, $b$, $c$) denotes model ** with $a$ transformer layer, $b$ body-part encoding dimension, and total $c$ parameter counts.
    We observe a rise-then-fall performance trend across three types of tasks as the model size increases.
    $\colorbox{red!15}{\rm Red background indicates best results}$.}
    \label{tab:ablation}
    \vspace{-0.4cm}
\end{table*}
\vspace{-0.2cm}
\paragraph{Comparison on Speech-to-Gesture Generation.}

In Tab. \ref{tab:s2g_m2d_res_1}, we compared \OurMethod~with MCM~\citep{mcm}, Talkshow~\citep{talkshow}, and EMAGE~\citep{beats2}. Our model achieved good quality and diversity in both hand and whole-body motion generation and excelled in aligning with the rhythm of first-perspective speech. This is credited to our coarse-to-fine training strategy and the robust topology knowledge learned from the static and dynamic human topology graphs.
However, in expressions, \OurMethod-Mix~performs slightly worse than EMAGE and Talkshow. This arises from origin dataset limitations in HumanML3D and FineDance, where the face was filled with random or average expressions, confusing the first training stage that affects the following S2G generation.
Still, we find \OurMethod-Mix possesses a notable performance boost against \OurMethod-Basic, further confirming that \OurAttn~learned robust topology knowledge that can be generalized across different generation scenarios.
Qualitative results in Fig. \ref{fig:visual} clearly show that \OurMethod~can effectively follow the beats and generate reasonable gestures and lip movements.

\vspace{-0.2cm}
\paragraph{Comparison on Music-to-Dance Generation.}

\OurMethod~achieves performance comparable to the SOTA baselines, as shown in Tab. \ref{tab:s2g_m2d_res_1}. Both variants of our model perform well in diversity, attributed to the first stage of coarse text-to-motion generation training. This equips the model with extensive motion topology knowledge across various scenarios. 
However, \OurMethod-Mix has an increase in \textbf{FID} compared to \OurMethod-Basic. This is likely due to the FineDance dataset's lack of necessary text descriptions, leading to identical pseudo-captions for different segments of the same song during the first stage of training. 
This one-to-many generation mode confuses when the model incorporates corresponding music information for each segment in the second stage, attempting to learn many-to-many relationships. 
Qualitative results in Fig. \ref{fig:visual} show that \OurMethod~can generate natural dances according to the music beats.

\subsection{Ablation Study}

We conducted ablation explorations about the necessity of \OurAttn~design and scaling up influences in Tab .\ref{tab:ablation}.

\begin{figure}[]
    \centering
    \includegraphics[width=1\linewidth]{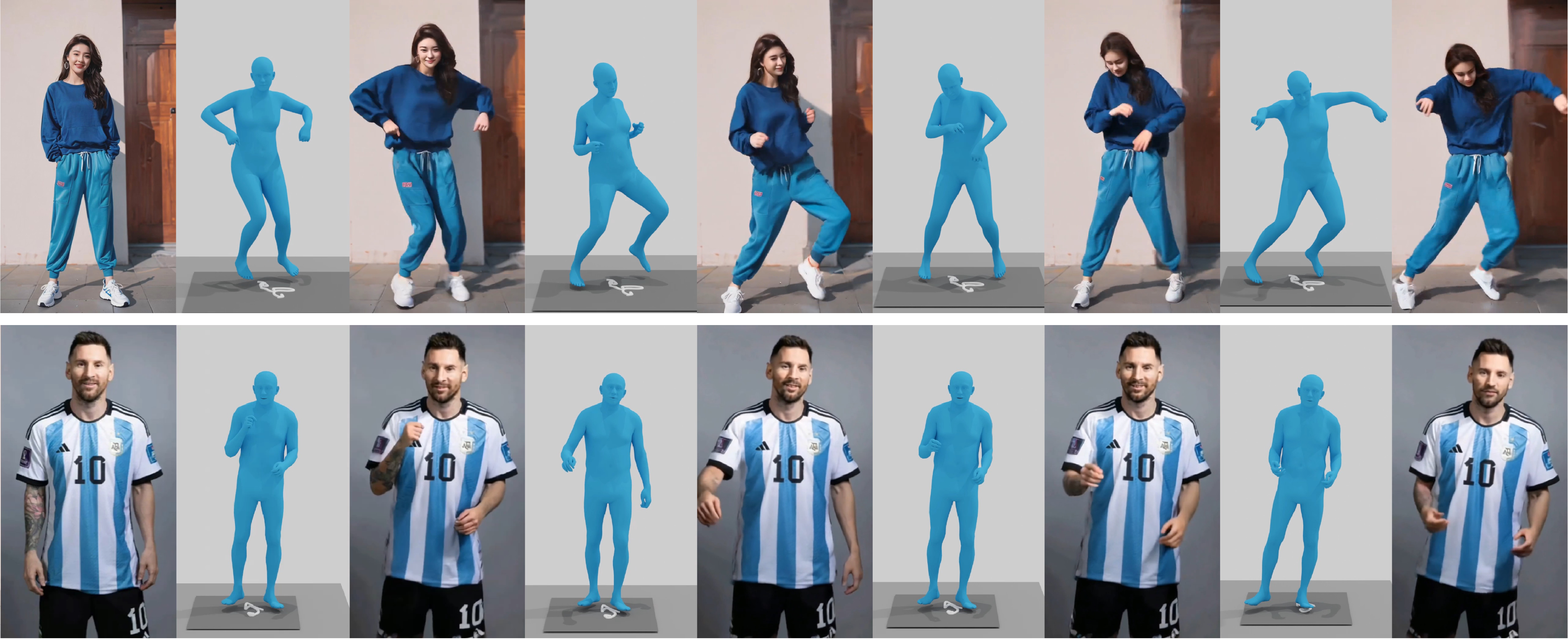}
    \vspace{-0.5cm}
    \caption{
    Multimodal video generation application with our generated motions conditioned on music~(upper row) or speech~(lower row). We project them to 2D images to serve as motion conditions for MimicMotion~\cite{mimicmotion}.
    }
    \vspace{-0.7cm}
    \label{fig:app}
\end{figure}

    \textbf{Different motion topology modeling designs.} We have three key observations about decoupling the static and dynamic human topology graph learning. 
    \textbf{\blackcircle{1}} 
    \textbf{\textit{Only modeling static topology decreases performance in T2M but significantly improves performance in S2G and M2D.} }
    We attribute this to the static topology ensuring the model grasps basic spatial relations between body parts, enhancing generalization across various generation scenarios. 
    However, the additional learnable spatial structure module, unrelated to input, increases learning difficulty in the T2M task. 
    \textbf{\blackcircle{2}} 
    \textbf{\textit{Only modeling dynamic topology nearly brings no benefit.}}
    This is because the initial optimization of the input-adaptive dynamic topology adjacency matrix is complex, especially for transferring topology knowledge against distribution drifts, making it hard to converge to the correct dynamic topology graph~\citep{lmm}. 
    \textbf{\blackcircle{3}}
    \textbf{\textit{Joint modeling of static and dynamic topologies effectively captures motion knowledge against distribution drifts}}, as in human-centric research~\citep{zeng2021learning}. The static topology learns basic human structure, providing foundational spatial knowledge across tasks, while the dynamic topology adjusts according to specific motion distributions and control signals.

    \textbf{Scaling up impacts.} Based on the acknowledgment of the scalability of transformer models, we explored the impact of model size on task performance. 
    We increased the size of \OurMethod-Mix~from $77M$ to $478M$, observing a rise-then-fall performance trend across three types of tasks as the model size increased with limited data. 
    This verifies that increasing the model's parameter size can enhance generative capabilities, but without a corresponding increase in high-quality data, model performance may decline.

\subsection{Application: Multimodal Video Generation}

To demonstrate the downstream application, in Fig.~\ref{fig:app}, we present two animation videos driven by \OurMethod~in M2D and S2G. Our generated motion sequences can be combined with any off-the-shelf human video generation framework, such as MimicMotion~\citep{mimicmotion}, AnimateAnyone~\citep{animateanyone}, and VividPose~\citep{vividpose}, enabling users to customize videos of any character based on specific control signals, such as speech or music. 
Notably, unlike the traditional 2D keypoints estimated from videos, our generated 3D motion approach allows for flexible adjustment of camera parameters to project different visible body regions (e.g., full body or upper body, as in Fig. \ref{fig:app}).
More detailed visualizations are in our supplementary. 
\section{Conclusion}
\label{sec:conclusion}

In this paper, we proposed \OurMethod, a unified framework for whole-body human motion generation with plug-and-play multimodal controls that generalizes across different generative distributions and efficiently handles control signals of varying granularity. 
\OurMethod~employs a coarse-to-fine training strategy that achieves fine-grained, plug-and-play control for different conditions~(including text, speech and music) without the optimization burden of mixed training.
Our core design is \OurAttn, which effectively learns and transfers motion knowledge across different distributions by parallel modeling the static and dynamic human topology graphs.
We introduced \Benchmark, the first available multimodal whole-body motion generation benchmark based on the unified whole-body SMPL-X representation.
Extensive experiments show that \OurMethod~achieves a competitive performance on standard motion generation tasks against current state-of-the-art baselines.

\bibliography{aaai25}

\begin{thebibliography}{55}
\providecommand{\natexlab}[1]{#1}

\bibitem[{Bian et~al.(2024)Bian, Ju, Li, Xu, Cheng, and Xu}]{bianmulti}
Bian, Y.; Ju, X.; Li, J.; Xu, Z.; Cheng, D.; and Xu, Q. 2024.
\newblock Multi-Patch Prediction: Adapting Language Models for Time Series Representation Learning.
\newblock In \emph{Forty-first International Conference on Machine Learning}.

\bibitem[{Chen et~al.(2024)Chen, Liu, Wang, Zeng, Li, and Chen}]{diffsheg}
Chen, J.; Liu, Y.; Wang, J.; Zeng, A.; Li, Y.; and Chen, Q. 2024.
\newblock DiffSHEG: A Diffusion-Based Approach for Real-Time Speech-driven Holistic 3D Expression and Gesture Generation.
\newblock In \emph{CVPR}.

\bibitem[{Chen et~al.(2023)Chen, Jiang, Liu, Huang, Fu, Chen, and Yu}]{mld}
Chen, X.; Jiang, B.; Liu, W.; Huang, Z.; Fu, B.; Chen, T.; and Yu, G. 2023.
\newblock Executing your Commands via Motion Diffusion in Latent Space.
\newblock In \emph{Proceedings of the IEEE/CVF Conference on Computer Vision and Pattern Recognition}, 18000--18010.

\bibitem[{Chung et~al.(2021)Chung, Wuu, Yang, Tai, and Tang}]{chung2021haa500}
Chung, J.; Wuu, C.-h.; Yang, H.-r.; Tai, Y.-W.; and Tang, C.-K. 2021.
\newblock Haa500: Human-centric atomic action dataset with curated videos.
\newblock In \emph{Proceedings of the IEEE/CVF international conference on computer vision}, 13465--13474.

\bibitem[{G{\"a}rtner et~al.(2022)G{\"a}rtner, Andriluka, Coumans, and Sminchisescu}]{gartner2022differentiable}
G{\"a}rtner, E.; Andriluka, M.; Coumans, E.; and Sminchisescu, C. 2022.
\newblock Differentiable dynamics for articulated 3d human motion reconstruction.
\newblock In \emph{Proceedings of the IEEE/CVF conference on computer vision and pattern recognition}, 13190--13200.

\bibitem[{Girdhar et~al.(2023)Girdhar, El-Nouby, Liu, Singh, Alwala, Joulin, and Misra}]{girdhar2023imagebind}
Girdhar, R.; El-Nouby, A.; Liu, Z.; Singh, M.; Alwala, K.~V.; Joulin, A.; and Misra, I. 2023.
\newblock Imagebind: One embedding space to bind them all.
\newblock In \emph{Proceedings of the IEEE/CVF Conference on Computer Vision and Pattern Recognition}, 15180--15190.

\bibitem[{Guo et~al.(2022)Guo, Zou, Zuo, Wang, Ji, Li, and Cheng}]{humanml3d}
Guo, C.; Zou, S.; Zuo, X.; Wang, S.; Ji, W.; Li, X.; and Cheng, L. 2022.
\newblock Generating Diverse and Natural 3D Human Motions From Text.
\newblock In \emph{Proceedings of the IEEE/CVF Conference on Computer Vision and Pattern Recognition (CVPR)}, 5152--5161.

\bibitem[{Guo et~al.(2020)Guo, Zuo, Wang, Zou, Sun, Deng, Gong, and Cheng}]{guo2020action2motion}
Guo, C.; Zuo, X.; Wang, S.; Zou, S.; Sun, Q.; Deng, A.; Gong, M.; and Cheng, L. 2020.
\newblock Action2motion: Conditioned generation of 3d human motions.
\newblock In \emph{Proceedings of the 28th ACM International Conference on Multimedia}, 2021--2029.

\bibitem[{Hu(2024)}]{hu2024animate}
Hu, L. 2024.
\newblock Animate anyone: Consistent and controllable image-to-video synthesis for character animation.
\newblock In \emph{Proceedings of the IEEE/CVF Conference on Computer Vision and Pattern Recognition}, 8153--8163.

\bibitem[{Hu et~al.(2023)Hu, Gao, Zhang, Sun, Zhang, and Bo}]{animateanyone}
Hu, L.; Gao, X.; Zhang, P.; Sun, K.; Zhang, B.; and Bo, L. 2023.
\newblock Animate Anyone: Consistent and Controllable Image-to-Video Synthesis for Character Animation.
\newblock \emph{arXiv preprint arXiv:2311.17117}.

\bibitem[{Kim, Kim, and Choi(2023)}]{kim2023flame}
Kim, J.; Kim, J.; and Choi, S. 2023.
\newblock Flame: Free-form language-based motion synthesis \& editing.
\newblock In \emph{Proceedings of the AAAI Conference on Artificial Intelligence}.

\bibitem[{Li et~al.(2021)Li, Yang, Ross, and Kanazawa}]{aistpp}
Li, R.; Yang, S.; Ross, D.~A.; and Kanazawa, A. 2021.
\newblock Learn to Dance with AIST++: Music Conditioned 3D Dance Generation.
\newblock arXiv:2101.08779.

\bibitem[{Li et~al.(2023)Li, Zhao, Zhang, Su, Ren, Zhang, Tang, and Li}]{finedance}
Li, R.; Zhao, J.; Zhang, Y.; Su, M.; Ren, Z.; Zhang, H.; Tang, Y.; and Li, X. 2023.
\newblock FineDance: A Fine-grained Choreography Dataset for 3D Full Body Dance Generation.
\newblock In \emph{Proceedings of the IEEE/CVF International Conference on Computer Vision}, 10234--10243.

\bibitem[{Liang et~al.(2024)Liang, Bao, Zhang, Ren, Xu, Yang, Chen, Yu, and Xu}]{liang2024omg}
Liang, H.; Bao, J.; Zhang, R.; Ren, S.; Xu, Y.; Yang, S.; Chen, X.; Yu, J.; and Xu, L. 2024.
\newblock Omg: Towards open-vocabulary motion generation via mixture of controllers.
\newblock In \emph{Proceedings of the IEEE/CVF Conference on Computer Vision and Pattern Recognition}, 482--493.

\bibitem[{Lin et~al.(2023{\natexlab{a}})Lin, Zeng, Lu, Cai, Zhang, Wang, and Zhang}]{motionx}
Lin, J.; Zeng, A.; Lu, S.; Cai, Y.; Zhang, R.; Wang, H.; and Zhang, L. 2023{\natexlab{a}}.
\newblock Motion-X: A Large-scale 3D Expressive Whole-body Human Motion Dataset.
\newblock \emph{Advances in Neural Information Processing Systems}.

\bibitem[{Lin et~al.(2023{\natexlab{b}})Lin, Zeng, Wang, Zhang, and Li}]{osx}
Lin, J.; Zeng, A.; Wang, H.; Zhang, L.; and Li, Y. 2023{\natexlab{b}}.
\newblock One-Stage 3D Whole-Body Mesh Recovery with Component Aware Transformer.
\newblock In \emph{CVPR}, 21159--21168.

\bibitem[{Ling et~al.(2023)Ling, Han, Wong, Kangkanhalli, and Geng}]{mcm}
Ling, Z.; Han, B.; Wong, Y.; Kangkanhalli, M.; and Geng, W. 2023.
\newblock Mcm: Multi-condition motion synthesis framework for multi-scenario.
\newblock \emph{arXiv preprint arXiv:2309.03031}.

\bibitem[{Liu et~al.(2024{\natexlab{a}})Liu, Zhu, Becherini, Peng, Su, Zhou, Zhe, Iwamoto, Zheng, and Black}]{beats2}
Liu, H.; Zhu, Z.; Becherini, G.; Peng, Y.; Su, M.; Zhou, Y.; Zhe, X.; Iwamoto, N.; Zheng, B.; and Black, M.~J. 2024{\natexlab{a}}.
\newblock EMAGE: Towards Unified Holistic Co-Speech Gesture Generation via Expressive Masked Audio Gesture Modeling.
\newblock In \emph{Proceedings of the IEEE/CVF Conference on Computer Vision and Pattern Recognition}, 1144--1154.

\bibitem[{Liu et~al.(2022{\natexlab{a}})Liu, Zhu, Iwamoto, Peng, Li, Zhou, Bozkurt, and Zheng}]{beats}
Liu, H.; Zhu, Z.; Iwamoto, N.; Peng, Y.; Li, Z.; Zhou, Y.; Bozkurt, E.; and Zheng, B. 2022{\natexlab{a}}.
\newblock Beat: A large-scale semantic and emotional multi-modal dataset for conversational gestures synthesis.
\newblock In \emph{European conference on computer vision}, 612--630. Springer.

\bibitem[{Liu et~al.(2023)Liu, Dai, Wang, Cheng, Tang, and Tong}]{liu2023plan}
Liu, J.; Dai, W.; Wang, C.; Cheng, Y.; Tang, Y.; and Tong, X. 2023.
\newblock Plan, posture and go: Towards open-world text-to-motion generation.
\newblock \emph{arXiv preprint arXiv:2312.14828}.

\bibitem[{Liu et~al.(2022{\natexlab{b}})Liu, Wu, Zhou, Xu, Qian, Lin, Zhou, Wu, Dai, and Zhou}]{liu2022learning}
Liu, X.; Wu, Q.; Zhou, H.; Xu, Y.; Qian, R.; Lin, X.; Zhou, X.; Wu, W.; Dai, B.; and Zhou, B. 2022{\natexlab{b}}.
\newblock Learning hierarchical cross-modal association for co-speech gesture generation.
\newblock In \emph{Proceedings of the IEEE/CVF Conference on Computer Vision and Pattern Recognition}, 10462--10472.

\bibitem[{Liu et~al.(2024{\natexlab{b}})Liu, Zhang, Li, Yan, Gao, Chen, Yuan, Huang, Sun, Gao, He, and Sun}]{sora}
Liu, Y.; Zhang, K.; Li, Y.; Yan, Z.; Gao, C.; Chen, R.; Yuan, Z.; Huang, Y.; Sun, H.; Gao, J.; He, L.; and Sun, L. 2024{\natexlab{b}}.
\newblock Sora: A Review on Background, Technology, Limitations, and Opportunities of Large Vision Models.
\newblock arXiv:2402.17177.

\bibitem[{Loper et~al.(2015)Loper, Mahmood, Romero, Pons-Moll, and Black}]{smpl}
Loper, M.; Mahmood, N.; Romero, J.; Pons-Moll, G.; and Black, M.~J. 2015.
\newblock {SMPL}: A Skinned Multi-Person Linear Model.
\newblock \emph{ACM Trans. Graphics (Proc. SIGGRAPH Asia)}, 34(6): 248:1--248:16.

\bibitem[{Lu et~al.(2023)Lu, Chen, Zeng, Lin, Zhang, Zhang, and Shum}]{humantomato}
Lu, S.; Chen, L.-H.; Zeng, A.; Lin, J.; Zhang, R.; Zhang, L.; and Shum, H.-Y. 2023.
\newblock HumanTOMATO: Text-aligned Whole-body Motion Generation.
\newblock \emph{arxiv:2310.12978}.

\bibitem[{Luo et~al.(2024)Luo, Hou, Chang, Liu, Wang, and Shan}]{m3gpt}
Luo, M.; Hou, R.; Chang, H.; Liu, Z.; Wang, Y.; and Shan, S. 2024.
\newblock M\^{3}-GPT: An Advanced Multimodal, Multitask Framework for Motion Comprehension and Generation.
\newblock \emph{arXiv preprint arXiv:2405.16273}.

\bibitem[{Ma, Bai, and Zhou(2022)}]{ma2022pretrained}
Ma, J.; Bai, S.; and Zhou, C. 2022.
\newblock Pretrained diffusion models for unified human motion synthesis. arXiv 2022.
\newblock \emph{arXiv preprint arXiv:2212.02837}.

\bibitem[{Mahmood et~al.(2019)Mahmood, Ghorbani, Troje, Pons-Moll, and Black}]{mahmood2019amass}
Mahmood, N.; Ghorbani, N.; Troje, N.~F.; Pons-Moll, G.; and Black, M.~J. 2019.
\newblock AMASS: Archive of motion capture as surface shapes.
\newblock In \emph{Proceedings of the IEEE/CVF international conference on computer vision}, 5442--5451.

\bibitem[{McFee et~al.(2015)McFee, Raffel, Liang, Ellis, McVicar, Battenberg, and Nieto}]{mcfee2015librosa}
McFee, B.; Raffel, C.; Liang, D.; Ellis, D.~P.; McVicar, M.; Battenberg, E.; and Nieto, O. 2015.
\newblock librosa: Audio and music signal analysis in python.
\newblock In \emph{SciPy}, 18--24.

\bibitem[{Nie et~al.(2022)Nie, Nguyen, Sinthong, and Kalagnanam}]{nie2022time}
Nie, Y.; Nguyen, N.~H.; Sinthong, P.; and Kalagnanam, J. 2022.
\newblock A time series is worth 64 words: Long-term forecasting with transformers.
\newblock \emph{arXiv preprint arXiv:2211.14730}.

\bibitem[{Oord, Li, and Vinyals(2018)}]{oord2018representation}
Oord, A. v.~d.; Li, Y.; and Vinyals, O. 2018.
\newblock Representation learning with contrastive predictive coding.
\newblock \emph{arXiv preprint arXiv:1807.03748}.

\bibitem[{Pavlakos et~al.(2019)Pavlakos, Choutas, Ghorbani, Bolkart, Osman, Tzionas, and Black}]{smplx}
Pavlakos, G.; Choutas, V.; Ghorbani, N.; Bolkart, T.; Osman, A. A.~A.; Tzionas, D.; and Black, M.~J. 2019.
\newblock Expressive Body Capture: {3D} Hands, Face, and Body from a Single Image.
\newblock In \emph{Proceedings IEEE Conf. on Computer Vision and Pattern Recognition (CVPR)}, 10975--10985.

\bibitem[{Petrovich, Black, and Varol(2022)}]{petrovich2022temos}
Petrovich, M.; Black, M.~J.; and Varol, G. 2022.
\newblock TEMOS: Generating diverse human motions from textual descriptions.
\newblock In \emph{European Conference on Computer Vision}, 480--497. Springer.

\bibitem[{Rombach et~al.(2022)Rombach, Blattmann, Lorenz, Esser, and Ommer}]{sd}
Rombach, R.; Blattmann, A.; Lorenz, D.; Esser, P.; and Ommer, B. 2022.
\newblock High-resolution image synthesis with latent diffusion models.
\newblock In \emph{Proceedings of the IEEE/CVF conference on computer vision and pattern recognition}, 10684--10695.

\bibitem[{Shazeer et~al.(2017)Shazeer, Mirhoseini, Maziarz, Davis, Le, Hinton, and Dean}]{moe}
Shazeer, N.; Mirhoseini, A.; Maziarz, K.; Davis, A.; Le, Q.; Hinton, G.; and Dean, J. 2017.
\newblock Outrageously large neural networks: The sparsely-gated mixture-of-experts layer.
\newblock \emph{arXiv preprint arXiv:1701.06538}.

\bibitem[{Siyao et~al.(2022)Siyao, Yu, Gu, Lin, Wang, Qian, Loy, and Liu}]{bailando}
Siyao, L.; Yu, W.; Gu, T.; Lin, C.; Wang, Q.; Qian, C.; Loy, C.~C.; and Liu, Z. 2022.
\newblock Bailando: 3d dance generation by actor-critic gpt with choreographic memory.
\newblock In \emph{Proceedings of the IEEE/CVF Conference on Computer Vision and Pattern Recognition}, 11050--11059.

\bibitem[{Tang et~al.(2023)Tang, Liu, Liu, Yang, Dai, Rao, Lu, Zhou, and Li}]{flag3d_cvpr}
Tang, Y.; Liu, J.; Liu, A.; Yang, B.; Dai, W.; Rao, Y.; Lu, J.; Zhou, J.; and Li, X. 2023.
\newblock FLAG3D: A 3D Fitness Activity Dataset with Language Instruction.
\newblock In \emph{CVPR}.

\bibitem[{Team et~al.(2023)Team, Anil, Borgeaud, Wu, Alayrac, Yu, Soricut, Schalkwyk, Dai, Hauth et~al.}]{team2023gemini}
Team, G.; Anil, R.; Borgeaud, S.; Wu, Y.; Alayrac, J.-B.; Yu, J.; Soricut, R.; Schalkwyk, J.; Dai, A.~M.; Hauth, A.; et~al. 2023.
\newblock Gemini: a family of highly capable multimodal models.
\newblock \emph{arXiv preprint arXiv:2312.11805}.

\bibitem[{Tevet et~al.(2023)Tevet, Raab, Gordon, Shafir, Cohen-or, and Bermano}]{mdm}
Tevet, G.; Raab, S.; Gordon, B.; Shafir, Y.; Cohen-or, D.; and Bermano, A.~H. 2023.
\newblock Human Motion Diffusion Model.
\newblock In \emph{The Eleventh International Conference on Learning Representations}.

\bibitem[{Trivedi, Thatipelli, and Sarvadevabhatla(2021)}]{trivedi2021ntu}
Trivedi, N.; Thatipelli, A.; and Sarvadevabhatla, R.~K. 2021.
\newblock NTU-X: an enhanced large-scale dataset for improving pose-based recognition of subtle human actions.
\newblock In \emph{Proceedings of the Twelfth Indian Conference on Computer Vision, Graphics and Image Processing}, 1--9.

\bibitem[{Tseng, Castellon, and Liu(2023)}]{edge}
Tseng, J.; Castellon, R.; and Liu, K. 2023.
\newblock Edge: Editable dance generation from music.
\newblock In \emph{Proceedings of the IEEE/CVF Conference on Computer Vision and Pattern Recognition}, 448--458.

\bibitem[{Vaswani et~al.(2017)Vaswani, Shazeer, Parmar, Uszkoreit, Jones, Gomez, Kaiser, and Polosukhin}]{vaswani2017attention}
Vaswani, A.; Shazeer, N.; Parmar, N.; Uszkoreit, J.; Jones, L.; Gomez, A.~N.; Kaiser, {\L}.; and Polosukhin, I. 2017.
\newblock Attention is all you need.
\newblock \emph{Advances in neural information processing systems}, 30.

\bibitem[{Wang et~al.(2024)Wang, Jiang, Xu, Zhang, Wang, Zhang, Cao, Cao, Wang, and Fu}]{vividpose}
Wang, Q.; Jiang, Z.; Xu, C.; Zhang, J.; Wang, Y.; Zhang, X.; Cao, Y.; Cao, W.; Wang, C.; and Fu, Y. 2024.
\newblock VividPose: Advancing Stable Video Diffusion for Realistic Human Image Animation.
\newblock \emph{arXiv preprint arXiv:2405.18156v1}.

\bibitem[{Xia et~al.(2020)Xia, Liu, Han, Wang, Gong, Liu, Niu, Tao, and Sugiyama}]{partnoise}
Xia, X.; Liu, T.; Han, B.; Wang, N.; Gong, M.; Liu, H.; Niu, G.; Tao, D.; and Sugiyama, M. 2020.
\newblock Part-dependent label noise: Towards instance-dependent label noise.
\newblock \emph{NeurIPS}, 33: 7597--7610.

\bibitem[{Yi et~al.(2023)Yi, Liang, Liu, Cao, Wen, Bolkart, Tao, and Black}]{talkshow}
Yi, H.; Liang, H.; Liu, Y.; Cao, Q.; Wen, Y.; Bolkart, T.; Tao, D.; and Black, M.~J. 2023.
\newblock Generating Holistic 3D Human Motion from Speech.
\newblock In \emph{CVPR}.

\bibitem[{Zeng et~al.(2021)Zeng, Sun, Yang, Zhao, Liu, and Xu}]{zeng2021learning}
Zeng, A.; Sun, X.; Yang, L.; Zhao, N.; Liu, M.; and Xu, Q. 2021.
\newblock Learning skeletal graph neural networks for hard 3d pose estimation.
\newblock In \emph{Proceedings of the IEEE/CVF international conference on computer vision}, 11436--11445.

\bibitem[{Zhang et~al.(2023{\natexlab{a}})Zhang, Yan, Xu, Feng, and Liew}]{magicavatar}
Zhang, J.; Yan, H.; Xu, Z.; Feng, J.; and Liew, J.~H. 2023{\natexlab{a}}.
\newblock MagicAvatar: Multi-modal Avatar Generation and Animation.
\newblock In \emph{arXiv:2308.14748}.

\bibitem[{Zhang et~al.(2023{\natexlab{b}})Zhang, Zhang, Cun, Huang, Zhang, Zhao, Lu, and Shen}]{t2mgpt}
Zhang, J.; Zhang, Y.; Cun, X.; Huang, S.; Zhang, Y.; Zhao, H.; Lu, H.; and Shen, X. 2023{\natexlab{b}}.
\newblock T2M-GPT: Generating Human Motion from Textual Descriptions with Discrete Representations.
\newblock In \emph{Proceedings of the IEEE/CVF Conference on Computer Vision and Pattern Recognition (CVPR)}.

\bibitem[{Zhang, Rao, and Agrawala(2023)}]{zhang2023adding}
Zhang, L.; Rao, A.; and Agrawala, M. 2023.
\newblock Adding conditional control to text-to-image diffusion models.
\newblock In \emph{Proceedings of the IEEE/CVF International Conference on Computer Vision}, 3836--3847.

\bibitem[{Zhang et~al.(2024{\natexlab{a}})Zhang, Cai, Pan, Hong, Guo, Yang, and Liu}]{motiondiffuse}
Zhang, M.; Cai, Z.; Pan, L.; Hong, F.; Guo, X.; Yang, L.; and Liu, Z. 2024{\natexlab{a}}.
\newblock Motiondiffuse: Text-driven human motion generation with diffusion model.
\newblock \emph{IEEE Transactions on Pattern Analysis and Machine Intelligence}.

\bibitem[{Zhang et~al.(2024{\natexlab{b}})Zhang, Jin, Gu, Hong, Cai, Huang, Zhang, Guo, Yang, He et~al.}]{lmm}
Zhang, M.; Jin, D.; Gu, C.; Hong, F.; Cai, Z.; Huang, J.; Zhang, C.; Guo, X.; Yang, L.; He, Y.; et~al. 2024{\natexlab{b}}.
\newblock Large motion model for unified multi-modal motion generation.
\newblock \emph{arXiv preprint arXiv:2404.01284}.

\bibitem[{Zhang et~al.(2023{\natexlab{c}})Zhang, Li, Cai, Ren, Yang, and Liu}]{finemogen}
Zhang, M.; Li, H.; Cai, Z.; Ren, J.; Yang, L.; and Liu, Z. 2023{\natexlab{c}}.
\newblock FineMoGen: Fine-Grained Spatio-Temporal Motion Generation and Editing.
\newblock \emph{NeurIPS}.

\bibitem[{Zhang et~al.(2022)Zhang, Ma, Zhang, Qian, Kwon, Pollefeys, Bogo, and Tang}]{zhang2022egobody}
Zhang, S.; Ma, Q.; Zhang, Y.; Qian, Z.; Kwon, T.; Pollefeys, M.; Bogo, F.; and Tang, S. 2022.
\newblock Egobody: Human body shape and motion of interacting people from head-mounted devices.
\newblock In \emph{European conference on computer vision}, 180--200. Springer.

\bibitem[{Zhang et~al.(2024{\natexlab{c}})Zhang, Gu, Wang, Wang, Cheng, Zhu, and Zou}]{mimicmotion}
Zhang, Y.; Gu, J.; Wang, L.-W.; Wang, H.; Cheng, J.; Zhu, Y.; and Zou, F. 2024{\natexlab{c}}.
\newblock MimicMotion: High-Quality Human Motion Video Generation with Confidence-aware Pose Guidance.
\newblock \emph{arXiv preprint arXiv:2406.19680}.

\bibitem[{Zhou et~al.(2019)Zhou, Barnes, Lu, Yang, and Li}]{zhou2019continuity}
Zhou, Y.; Barnes, C.; Lu, J.; Yang, J.; and Li, H. 2019.
\newblock On the continuity of rotation representations in neural networks.
\newblock In \emph{Proceedings of the IEEE/CVF conference on computer vision and pattern recognition}, 5745--5753.

\bibitem[{Zhou, Wan, and Wang(2023)}]{ude2}
Zhou, Z.; Wan, Y.; and Wang, B. 2023.
\newblock A unified framework for multimodal, multi-part human motion synthesis.
\newblock \emph{arXiv preprint arXiv:2311.16471}.

\end{thebibliography}

\newpage
\section{Appendix}
\subsection{Visualization Results}

Due to the limitations of the PDF's static format and the page limit, additional visualizations and comparisons are available in \textbf{the supplementary and project page}, which include \textbf{\textit{generation visualizations for individual tasks}} such as Text-to-Motion~(T2M), Music-to-Dance~(M2D), and Speech-to-Gesture~(S2G), as well as \textbf{\textit{visualizations of plug-and-play control generation within a single long sequence}}. The video also demonstrates the application of our generated motion sequences in \textbf{\textit{video production and character animation}}.

\subsection{Related Works}
\subsubsection{Human Motion Generation Models}
Conditioned human motion generation models have made significant progress, including text-to-motion~(T2M)~\citep{mdm,t2mgpt,liu2023plan,motiondiffuse,finemogen,liang2024omg}, speech-to-gesture~(S2G)~\citep{talkshow,diffsheg,liu2022learning}, and music-to-dance~(M2D)~\citep{finedance,edge,bailando}.
In text-to-motion, models~\citep{mdm,mld,t2mgpt,liu2023plan,motiondiffuse,finemogen,liang2024omg} achieve text-controlled motion generation with semantic consistency by applying advanced generative models and aligning motion and text feature domains. 
For speech-to-gesture~\citep{talkshow,diffsheg,liu2022learning}, many efforts focus on mapping speech to human gestures through rhythm alignment and character style learning. 
Additionally, numerous studies~\citep{finedance,edge,bailando} design spatial and temporal coherence constraints to ensure that models learn the corresponding style and rhythm from the input music.
Recently, increasing attention has been given to multimodal motion generation~\citep{mcm,lmm,m3gpt}. 
%
%
%
$M^{3}$-GPT~\citep{m3gpt} injects quantized condition tokens into the vocabulary of large language models to achieve motion understanding and generation, but it overlooks the modeling of human topology priors.
Motion-Verse~\citep{lmm} incorporates dynamic attention to assess relationships among body parts but fails to capture the overall static human topology, leading to limited generalization and increased optimization complexity. 
Furthermore, it employs mixed training across all conditions based on ImageBind~\citep{girdhar2023imagebind}, which creates optimization challenges when learning conditions of varying granularity simultaneously and needs retraining for new control signals.
%
MCM~\citep{mcm} attempts to address the optimization confusion of mixed training based on the ControlNet~\citep{zhang2023adding} architecture, but it neglects any modeling of human topology structure, resulting in poor generalization across generation scenarios. 
Moreover, MCM only focuses on tensor movements, lacking the ability to generate whole-body motion.
Compared to previous methods in Tab.~\ref{tab:compare_other_methods}, \OurMethod~generates whole-body motion under varying control signals with plug-and-play capability by using \OurAttn~to capture static human topology and domain-specific dynamic skeleton relationships, incorporating control branches, and employing a coarse-to-fine training strategy.

\subsubsection{Human Motion Generation Benchmarks}
Various conditioned human motion generation benchmarks have been constructed in recent years.
For T2M, researchers have curated datasets encompassing action categories~\citep{chung2021haa500,trivedi2021ntu}, sequential action labels~\citep{zhang2022egobody,guo2020action2motion}, and arbitrary natural language descriptions~\citep{motionx,humanml3d,flag3d_cvpr} at various abstraction levels. 
Specifically, AMASS~\citep{mahmood2019amass} consolidates 15 optical marker-based motion capture datasets into a comprehensive collection based on SMPL~\citep{smpl} representation. 
HumanML3D~\citep{humanml3d} extracted a high-quality subset within AMASS~\citep{mahmood2019amass} based on H3D format for torso-only generation, including three arbitrary natural language descriptions per motion clip from diverse annotators.
For M2D, 
AIST++~\citep{aistpp} reconstructs 5 hours of dance based on SMPL~\citep{smpl} format from videos, despite its significant reconstruction error, and lack of capture of hand movements. 
Finedance~\citep{finedance} collects dances of 14.6 hours across 22 genres and supplements the dataset with detailed gestures using the SMPL-H~\citep{smplx} format.
For S2G, datasets~\citep{beats2,beats,talkshow} are gathered from pseudo-labeled~(PGT) and motion-captured sources. Mocap datasets are generally preferred due to significant errors in monocular 3D pose estimation in PGT~\citep{gartner2022differentiable}. Recently,
BEAT2~\citep{beats2} and BEAT~\citep{beats} have emerged as the most popular benchmarks, celebrated for their diverse range of motion and extensive data volume. BEAT2, building upon BEAT, utilizes SMPL-X and FLAME~\citep{kim2023flame} to achieve higher-quality unified mesh-level data.
Despite these developments, no publicly available benchmark supports unified representation for multimodal whole-body motion generation.

\subsection{\Benchmark~Construction}
\subsubsection{Motion Representation}

From body-only motion generation~\citep{humanml3d,aistpp,mcm} to whole-body motion generation~\citep{humantomato,finedance,beats2,lmm}, previous research has explored various motion representations in generation tasks, including the default axis-angle input based on SMPL mesh parameters~\citep{smpl}, 6D rotation~\citep{finedance}, quaternion~\citep{smplx}, and the extended H3D-format from SMPL~\citep{humanml3d}, which adds redundant information like joint positions and velocities. In recent years, SMPL-X~\citep{smplx}, an extension of SMPL, has incorporated hand modeling to enable finer-grained finger joint modeling. 
Therefore, considering the practicality and efficiency of motion representation, we use the default axis-angle input of SMPL-X to model the main body and hands.
Specifically, the $i$-th pose is defined by a tuple of root axis-angle ($\dot{r}^r\in \mathbb{R}^{3}$) around the X(Y and Z)-axis, root trajectory ($\dot{r}^t\in \mathbb{R}^{3}$) along the X(Y and Z)-axis, local joints axis-angle rotations ($\mathbf{\theta}^r \in \mathbb{R}^{3N}$), where $N$ denotes the number of whole body joints, including both body joints and hand joints.
For face motion representations, we follow the MotionX~\citep{motionx} to adopt the $\mathbf{f}^{s} \in \mathbb{R}^{100}$ to represent the face shape, $\mathbf{f}^{e} \in \mathbb{R}^{50}$ in the Flame Format~\citep{kim2023flame} to represent the face expression, and jaw axis-angle rotation ($\mathbf{\theta}^j \in \mathbb{R}^{3}$) for jaw modeling.
Additionally, we employ the standard SMPL-X model 10-dimensional parameters $\mathbf{\theta}^b \in \mathbb{b}^{10}$ to represent the body shape.
Thus, we represent the whole-body motion as $\mathbf{m}_i = \{\dot{r}^r, \dot{r}^t, \mathbf{\theta}^r, \mathbf{f}^{s}, \mathbf{f}^{e}, \mathbf{\theta}^j, \mathbf{\theta}^b\}$.

\subsubsection{Text-to-Motion Subset Construction}
In the first phase of semantic text-to-motion pre-training, our data primarily consists of three parts, as follows:

\begin{enumerate}
    \item \textbf{HumanML3D}~\citep{humanml3d} is a representative 3D motion-text dataset containing 14,616 high-quality human motions paired with 44,970 text captions. Instead of using the body-only H3D format from the original HumanML3D, we re-extracted the corresponding instances from its original AMASS~\citep{mahmood2019amass} data in the SMPL-X format and processed each motion frame based on our SMPL-X axis-angle format, setting the corresponding SMPL-X~\citep{smplx} representation to zero for any missing body parts. The text was filtered and processed according to the original HumanML3D text caption processing workflow.

    \item \textbf{BEAT2}\citep{beats2} is a speech-to-gesture dataset that includes various speaking styles and speaker IDs. It provides SMPL-X~\citep{smplx} axis-angle rotation motion representation, from which we directly extract the corresponding rotation information based on our whole-body motion format $\mathbf{m}_i = \{\dot{r}^r, \dot{r}^t, \mathbf{\theta}^r, \mathbf{f}^{s}, \mathbf{f}^{e}, \mathbf{\theta}^j, \mathbf{\theta}^b\}$. For the text part, we generate corresponding pseudo-semantic text captions using simple rules, such as "A person is giving a speech, and the content is ...".

    \item \textbf{FineDance}\citep{finedance} is currently one of the leading music-to-dance datasets in terms of choreography diversity and data volume, originally providing body-hand data representation based on SMPL-H~\citep{smplx} rot6D. We first convert the simplified rot6D rotation matrix into the axis-angle format around the XYZ axis, leveraging the equivalence between different rotation representations~\citep{zhou2019continuity}. Instead of using the official SMPL-H~\citep{smplx} to SMPL-X~\citep{smplx} retargeting optimization method, we directly map SMPL-H parameters to SMPL-X~\citep{smplx}. Our qualitative and quantitative experiments demonstrate that this simple approach is effective, with negligible retargeting errors. For the textual part, we apply basic rules to generate pseudo-semantic captions matching the corresponding music segments, such as "A dancer is performing a street dance in the Jazz style to the rhythm of the wildfire."
\end{enumerate}

\subsubsection{Speech-to-Gesture Subset Construction}

For all speeches in BEAT2~\citep{beats2}, we use Librosa~\citep{mcfee2015librosa} to extract 2-dimensional temporal speech features related to speech prosody. The audio is sampled at 76,800 Hz with a hop size of 512. We segment the motion sequences and corresponding speech into 64-frame segments with a stride of 64 frames. In subsequent generations, we employ the outpainting-based sampling strategy from DiffSHEG~\citep{diffsheg} to achieve long-term gesture generation.
The pre-process details of motion representation and semantic text captioning have been thoroughly explained above, so they will not be repeated here.

\subsubsection{Music-to-Dance Subset Construction}

For all music in FineDance~\citep{finedance}, we use Librosa~\citep{mcfee2015librosa} to extract 35-dimensional temporal music features. The audio is sampled at 76,800 Hz with a hop size of 512. We segment the motion sequences and corresponding music into 120-frame segments with a stride size of 30 frames.
The pre-process details of motion representation and semantic text captioning have been thoroughly explained above, so they will not be repeated here.

\begin{table*}[!ht]
    \centering
    \resizebox{1.0\linewidth}{!}{
        \begin{tabular}{c|ccc|ccc}
            \toprule
            \multirow{2}{*}{Method} & \multicolumn{3}{c|}{R Precision} & \multirow{2}{*}{FID $\downarrow$} & \multirow{2}{*}{Div $\uparrow$} & \multirow{2}{*}{MM Dist$\downarrow$}  \\
            \cline{2-4}
             & Top-1 $\uparrow$ & Top-2 $\uparrow$ & Top-3 $\uparrow$ \\
            \hline
            GT &$0.511^{\pm{0.003}}$ & $0.703^{\pm{0.003}}$ & $0.797^{\pm{0.002}}$ &  $0.002^{\pm{0.000}}$ & $9.503^{\pm{0.065}}$ & $2.974^{\pm{0.008}}$  \\
            \hline
            T2M-GPT\cite{t2mgpt} & $0.491^{\pm{0.003}}$ & $0.680^{\pm{0.003}}$ & $0.775^{\pm{0.002}}$ &  \cellcolor{yellow!15}$0.116^{\pm{0.004}}$ & \cellcolor{red!15}$9.761^{\pm{0.081}}$ & $3.118^{\pm{0.011}}$ \\
            MDM\cite{mdm} & $0.418^{\pm{0.005}}$ & $0.604^{\pm{0.005}}$ & $0.707^{\pm{0.004}}$ &  $0.489^{\pm{0.025}}$ & $9.450^{\pm{0.066}}$ & $3.630^{\pm{0.023}}$ \\
            MotionDiffuse\cite{motiondiffuse} & $0.491^{\pm{0.001}}$ & $0.681^{\pm{0.001}}$ & $0.782^{\pm{0.001}}$ &  $0.630^{\pm{0.001}}$ & $9.410^{\pm{0.049}}$ & $3.113^{\pm{0.001}}$ \\
            FineMoGen\cite{finemogen} & \cellcolor{red!15}$0.504^{\pm{0.002}}$ & \cellcolor{yellow!15}$0.690^{\pm{0.002}}$ & $0.784^{\pm{0.002}}$ &  $0.151^{\pm{0.008}}$ & $9.263^{\pm{0.094}}$ & \cellcolor{red!15}$2.998^{\pm{0.008}}$ \\
            Motion-Verse\cite{lmm} & $0.496^{\pm{0.002}}$ & $0.685^{\pm{0.002}}$ & \cellcolor{yellow!15}$0.785^{\pm{0.002}}$ &  $0.415^{\pm{0.002}}$ & $9.176^{\pm{0.074}}$ & $3.087^{\pm{0.012}}$ \\
            MCM\cite{mcm} & $0.494^{\pm{0.003}}$ & $0.682^{\pm{0.005}}$ & $0.777^{\pm{0.003}}$ &  \cellcolor{red!15}$0.075^{\pm{0.003}}$ & $9.484^{\pm{0.074}}$ & $3.086^{\pm{0.011}}$ \\
            
            \hline
            \OurMethod-Basic & \cellcolor{yellow!15}$0.501^{\pm{0.003}}$ & \cellcolor{red!15}$0.697^{\pm{0.003}}$ & \cellcolor{red!15}$0.796^{\pm{0.002}}$ & $0.173^{\pm{0.002}}$ & \cellcolor{yellow!15}$9.543^{\pm{0.098}}$ & \cellcolor{yellow!15}$3.025 ^{\pm{0.008}}$ \\
            \bottomrule
        \end{tabular}
    }
    \vspace{-0.2cm}
    \caption{\textbf{Results of text-to-motion in origin HumanML3D benchmark.} We compare the results of text-to-motion generation between ours and the SOTA methods. Our method achieves better semantic relevance, fidelity, and diversity performances.  $\colorbox{red!15}{\rm Red background indicates best results}, \colorbox{yellow!15}{\rm yellow background indicates second best results}$.}
    \label{tab:t2m_res_1}
\end{table*}
\begin{table*}
    \centering
    \resizebox{1.0\linewidth}{!}{
        \begin{tabular}{cc|ccccc|ccccc|ccc}
            \toprule
            \multicolumn{2}{c|}{Method} & \multicolumn{5}{c|}{HumanML3D~(Text-to-Motion)} & \multicolumn{5}{c|}
            {BEAT2~(Speech-to-Gesture)}  & \multicolumn{3}{c}{Finedance~(Music-to-Dance)} \\
            \midrule
            {Local-Unfreeze} & 
            {Temporal-Patching} & 
            Top-1 $\uparrow$ & Top-2 $\uparrow$ & Top-3 $\uparrow$ & {FID $\downarrow$} & {Div $\uparrow$} & {$FID_{H}$ $\downarrow$} & {$FID_{B}$ $\downarrow$} & Face L2 $\downarrow$ & {Beat Align Score $\uparrow$} & {Div $\uparrow$} & {$FID_{H}$ $\downarrow$} & {$FID_{B}$ $\downarrow$} & {Div $\uparrow$} \\
             \midrule
             \redcheck & \redcheck & \cellcolor{red!15}{0.653} & \cellcolor{red!15}{0.794} & \cellcolor{red!15}{0.847} & \cellcolor{red!15}{5.593} & \cellcolor{red!15}{36.264} & \cellcolor{red!15}{15.346} & 27.140 & \cellcolor{red!15}{8.322} & 8.023 & 11.024 &   \cellcolor{red!15}{2.370}  & {59.471}  & \cellcolor{red!15}17.036 \\
             \redcheck & \greencheck & 0.628 & 0.776 & 0.834 & 5.944 & 36.189 & 17.583 & 27.605 & 8.792 & 8.007 & 10.920 & 3.496  & 64.784 & 16.371 \\
             \greencheck & \redcheck & - & - & - & - & - & 17.962 & \cellcolor{red!15}{26.556} & 8.561 & \cellcolor{red!15}{8.035} & \cellcolor{red!15}{11.248} & 2.493  & \cellcolor{red!15}56.847 & 16.894 \\
             \greencheck & \greencheck & - & - & - & - & - & 18.554 & 28.434 & 8.630 & 7.980 & 11.157 &  3.229 & 61.518 & 16.502 \\
             \bottomrule
        \end{tabular}
    }
    \caption{\textbf{Additional Ablation Study.} 
    we explored the second stage model training strategy about the body-wise encoder~(decoder) and motion sequence temporal relationship modeling paradigm.   
    The “Local-Unfreeze” column indicates that during the second phase, only specific body parts corresponding to certain control signals are unfrozen in the body-wise encoder and decoder. For instance, in the Speech-to-Gesture task, only the encoders and decoders for hands and face are unfrozen, while in Music-to-Dance, only the encoder and decoder for the hands are unfrozen.
    The "Temporal-Patching" column means performing patching operations on adjacent frames, compressing a specified number of neighboring frames into a single basic unit on the time dimension for modeling, instead of treating each frame's motion as a basic unit on the time dimension.
    $\colorbox{red!15}{\rm Red background indicates best results}$.}
    \label{tab:ablation-2}
    \vspace{-0.5cm}
\end{table*}
\subsection{Experiments}
\subsubsection{Implementation Details of \OurMethod}

We employ a 4-layer motion diffusion transformer as the backbone of \OurMethod, featuring a latent dimension of $12 \times 64$ and a feedforward embedding size of $256$, where $12$ corresponds to the number of body parts and $64$ denotes the dimensionality of each body-specific hidden state. 
For the control branch of \OurMethod, we set the number of copied \OurMethod~blocks to 2, representing half of the total. The encoder design for various low-level control signals (speech or music) aligns with the baselines~\citep{beats2,finedance}.
For the text encoder, we utilize a frozen CLIP ViT-B/32 encoder, enhanced with two additional transformer encoder layers. In the diffusion model, the variances $\beta_t$ are predefined to linearly range from $0.0001$ to $0.02$, with 1000 noising steps. Following MDM~\citep{mdm}, we set $x_{start}$ as the diffusion prediction goal instead of the noise. The model is trained using the Adam optimizer, starting with a learning rate of $2\times10^{-4}$, which decays to $2\times10^{-5}$ via a cosine schedule. Training occurs on $8\times$ NVIDIA Tesla V100-32GB GPUs, with a batch size of $64$ per GPU, and takes approximately 48 hours.

\subsubsection{Implementation Details of Text-Motion Retrieval Pre-Training in Evaluation}

Since we used SMPL-X-based axis-angle as the motion representation, the motion encoder and text encoder from previous studies could not be directly applied for evaluation. Therefore, following HumanTomato~\cite{humantomato}, we retrained a text-whole-body-motion retrieval model specifically for our SMPL-X axis-angle motion representation to assess performance in a contrastive learning approach. 
This retrieval model employs a VAE-based architecture~\citep{petrovich2022temos} consisting of a motion encoder, a text encoder, and a motion decoder. The training objective is the weighted sum of:
\begin{equation*}
    \min \mathcal{L}_{rec} + \lambda_{KL}\mathcal{L}_{KL} + \lambda_{E}\mathcal{L}_{E} + \mathcal{\lambda}_{NCE}\mathcal{L}_{NCE},
\end{equation*}
where the four loss terms are reconstruction loss, Kullback-Leibler (KL) divergence loss, cross-modal embedding similarity loss, and InfoNCE~\citep{oord2018representation} loss, respectively. The hyperparameters are set to $\lambda_{KL}=1\times10^{-5}, \lambda_{E}=1\times10^{-5}, \lambda_{NCE}=1\times10^{-1}$.

\subsubsection{More Results on Text-to-Motion}

To provide a more comprehensive comparison, in addition to evaluating on the HumanML3D subset with whole-body format SMPL-X in \Benchmark~(Tab. \ref{tab:t2m_res_2}), we also compare \OurMethod~with current SOTA baselines~\citep{finemogen,mcm,lmm,motiondiffuse,mdm,t2mgpt} on the original HumanML3D~\citep{humanml3d} using the body-only H3D format, which contains redundant information. Quantitative comparison results are shown in Tab. \ref{tab:t2m_res_1}. It is evident that on the original HumanML3D text-to-motion benchmark, \OurMethod~also achieves better text-guided generation capability, diversity, and motion generation quality.
 
Notably, in the HumanML3D subset of \Benchmark, the limited evaluation capabilities observed in the original HumanML3D benchmark with torso-only representation—where performance differences between models were minimal—were significantly enhanced. This improvement arises because the whole-body SMPL-X representation necessitates the model to generate torso movements, gestures, and expressions, rather than focusing solely on the torso.

\subsubsection{More Results on Ablation Study}

In addition to the ablation experiments on dynamic-static motion topology modeling and model parameter scaling in Tab. \ref{tab:ablation}, we further explored the second stage model training strategy and motion sequence temporal relationship modeling paradigm. The results are in Tab. \ref{tab:ablation-2}.

\textbf{\textit{Ablation on the second stage body-wise encoder and decoder training strategy.}} The “Local-Unfreeze” column in Tab. \ref{tab:ablation-2} indicates that during the second phase, only specific body parts corresponding to certain control signals are unfrozen in the body-wise encoder and decoder. For instance, in the Speech-to-Gesture task, only the encoders and decoders for hands and face are unfrozen, while in Music-to-Dance, only the encoder and decoder for the hands are unfrozen. 
Rows one and three of Tab. \ref{tab:ablation-2} clearly show that fully unfreezing the body-wise encoder and decoder during the second phase enhances encoding and decoding optimization for specific body parts in the generation scenario, thus improving the generation capabilities for targeted scenarios (e.g., hand and face modeling in Speech-to-Gesture, hand modeling in Music-to-Dance). Conversely, partial unfreezing helps retain the human body topology knowledge learned during the first phase of text semantic pre-training, thereby stabilizing the overall generation capability for full-body actions on downstream generation tasks.

\textbf{\textit{Ablation on motion sequence temporal relationship modeling.}}
Currently, there are two classic approaches for modeling temporal relationships in motion sequences: treating each frame's motion as a basic unit on the time dimension for sequence modeling, and performing patching operations on adjacent frames, compressing a specified number of neighboring frames into a single basic unit on the time dimension for modeling. In general time series analysis, the latter method has been widely shown to significantly improve performance in transformer-based models~\citep{nie2022time}, as it mitigates the impact of extreme values and eliminates redundant information, allowing for higher information density in sub-sequence relationship modeling. However, as shown in Table \ref{tab:ablation-2}, the conclusions for temporal dynamic modeling of motion sequences appear to be the opposite of those for general time series modeling. This discrepancy can be attributed to differences in data representation between motion and general time series:
\begin{itemize}
    \item High-quality motion data~\citep{humanml3d,finedance,beats2} from motion capture systems generally do not suffer from extreme values, whereas general time series data~\citep{nie2022time}can be affected by various factors such as collection environment and economic or cultural influences, leading to inconsistent quality.
    \item In SMPL-X axis-angle motion format, the rotation angles of child joints are influenced by their parent joints, meaning small changes in parent joint values can lead to significant motion changes due to the whole-body topological structure~\citep{smpl,smplx}. In other words, the axis-angle motion representation in SMPL-X is far more sensitive than general time series data, and using compressed sub-sequences as modeling units can introduce significant cumulative errors.
\end{itemize}

\subsection{Broader Impact and Limitation}

In this section, we will discuss the possible social impact and limitations of \OurMethod.

\paragraph{Broader Impact.}

First, we explore the task of whole-body motion generation under multimodal controls and establish the first multimodal motion generation benchmark with a unified whole-body motion representation based on three high-quality single-control signal motion generation datasets. These could serve as a foundation for the multimodal control motion generation research community.
Additionally, with large-scale motion data training under multimodal controls, our trained \OurMethod~can function as a motion prior for other research, such as HumanTomato~\citep{humantomato} and VPoser~\citep{smplx}. It can also help address noisy annotations in the current Motion Capture process~\citep{partnoise,osx}.
Finally, expressive, multimodal-controllable, and high-quality motion generation can be applied to various practical downstream scenarios, including but not limited to human video generation, motion animations, and robotics.

\paragraph{Limitation.}

While this work achieves significant progress in whole-body motion generation under multimodal controls, some limitations remain. 
First, the utilization of more high-quality semantic text descriptions for whole-body motion generation requires further investigation. This work follows previous approaches by using sequential semantic descriptions without incorporating frame-level or fine-grained whole-body descriptions throughout the two-stage training. This issue is particularly important in the Speech-to-Gesture~\citep{beats2} and Music-to-Dance~\citep{finedance} tasks, where sequence-level semantic captions are naturally lacking, and pseudo-text descriptions are used to supplement them.
Second, the current motion representation employs the axis-angle SMPL-X format. Although this allows direct rendering of the corresponding mesh via the SMPL-X model~\citep{smplx}, the 6D parameters for root rotation and root trajectory can significantly affect the overall motion generation quality, causing additional fluctuations during model training. In the future, we will explore adding projected 3D joint positions, similar to the redundant H3D format in HumanML3D~\citep{humanml3d}, to provide extra constraints on root rotation and root trajectory.
Additionally, we plan to unify more multimodal datasets to advance the development of a superior multimodal whole-body motion generation model.
\end{document}